\documentclass[10pt,twocolumn,letterpaper]{article}

\usepackage{cvpr}              

\usepackage[utf8]{inputenc}  
\usepackage[T1]{fontenc}     
\usepackage[pagebackref,breaklinks,colorlinks]{hyperref}
\usepackage{url}             
\usepackage{amsfonts}        
\usepackage{amsmath,amssymb} 
\usepackage{nicefrac}        
\usepackage{microtype}       
\usepackage{xcolor}          
\usepackage{xspace}

\usepackage{epsfig}
\usepackage{graphicx}
\usepackage{wrapfig}

\usepackage{multirow}
\usepackage{rotating}
\usepackage{booktabs}
\usepackage{tabularx}
\usepackage{colortbl}

\usepackage{paralist}

\def\cvprPaperID{9621}
\def\confName{CVPR}
\def\confYear{2023}

\newcolumntype{s}{>{\columncolor[gray]{.85}[.5\tabcolsep]}c}

\definecolor{pltgreen}{RGB}{156,194,229}  

\newcommand{\tabref}[1]{Tab.~\ref{#1}\xspace}
\newcommand{\figref}[1]{Fig.~\ref{#1}\xspace}
\newcommand{\secref}[1]{Sec.~\ref{#1}\xspace}

\newcommand{\textquote}[1]{\textit{#1}}
\newcommand{\paradigmfull}{Iterative Vision-and-Language Navigation}
\newcommand{\paradigm}{IVLN}
\newcommand{\benchmarkfull}{Iterative Room-to-Room}
\newcommand{\benchmark}{IR2R}
\newcommand{\benchmarkce}{IR2R-CE}
\newcommand{\action}[1]{\texttt{#1}}
\newcommand{\model}[1]{\texttt{#1}}
\newcommand{\metric}[1]{\texttt{#1}}

\newcommand{\thamt}{TourHAMT\xspace}

\renewcommand{\paragraph}[1]{\noindent\textbf{#1}}

\begin{document}

\title{Iterative Vision-and-Language Navigation}

\author{
  Jacob Krantz$^1$\thanks{Equal contributions. Correspondence: \href{mailto:krantzja@oregonstate.edu}{krantzja@oregonstate.edu}} \qquad Shurjo Banerjee$^{2\phantom{|}*}$ \qquad Wang Zhu$^3$ \\ Jason Corso$^2$ \qquad Peter Anderson$^4$ \qquad Stefan Lee$^1$ \qquad Jesse Thomason$^3$ \\
  {\normalsize $^1$Oregon State University} \quad
  {\normalsize $^2$University of Michigan} \quad
  {\normalsize $^3$University of Southern California} \quad
  {\normalsize $^4$Google Research}
}
\maketitle

\begin{abstract}
    We present \paradigmfull~(\paradigm), a paradigm for evaluating language-guided agents navigating in a persistent environment over time.
    Existing Vision-and-Language Navigation (VLN) benchmarks erase the agent's memory at the beginning of every episode, testing the ability to perform cold-start navigation with no prior information.
    However, deployed robots occupy the same environment for long periods of time.
    The \paradigm\ paradigm addresses this disparity by training and evaluating VLN agents that maintain memory across \textit{tours} of scenes that consist of up to 100 ordered instruction-following Room-to-Room (R2R) episodes, each defined by an individual language instruction and a target path.
    We present discrete and continuous \benchmarkfull\ (\benchmark) benchmarks comprising about 400 tours each in 80 indoor scenes.
    We find that extending the implicit memory of high-performing transformer VLN agents is not sufficient for \paradigm, but agents that build maps can benefit from environment persistence, motivating a renewed focus on map-building agents in VLN.
\end{abstract}

\section{Introduction}
\label{sec:intro}

Robots and virtual agents that persistently operate in human spaces like homes should improve over time.
For example, a smart vacuum told to \textquote{clean the living room, which is down the hall past the guest bedroom} should learn about both the living room and guest bedroom.
Likewise, agents should be able to associate references in past instructions, such as \textquote{guest bedroom}, with spatial and visual information from the environment to understand future instructions.

Most work on language-guided, embodied agents performing navigation~\cite{anderson:cvpr18,krantz_vlnce_2020} or household tasks~\cite{shridhar:cvpr20} is \textit{episodic} in nature---agent memory is erased before issuing each new instruction.
In contrast, physical robots build maps~\cite{durrant2006simultaneous,thrun2007simultaneous,zhang2017neural} \textit{iteratively} from visual observations~\cite{taketomi2017visual,merzlyakov2021comparison} as an explicit form of long-term memory.
Agents trained to perform language-guided navigation in simulation that are deployed on physical robots~\cite{anderson2020sim} fail to take advantage of the mapping-based strategies that facilitate robot navigation.

We propose \paradigmfull\ (\paradigm), in which an agent follows an \textit{ordered sequence} of language instructions that conduct a \textit{tour} of an indoor space.
Each tour is composed of individual \textit{episodes} of language instructions with target paths.
Agents can utilize \textit{memory} to better understand future tour instructions.
After just 10 episodes an agent has seen on average over 50\% of the target path associated with the next language instruction in a tour.
While performing an \paradigm\ tour, agents iteratively explore the environment, meaning regions irrelevant to task instructions need not ever be visited. 
By conditioning exploration on language, \paradigm\ enables rich semantic representations, \eg, unusual, novel, and scene-specific referents grounded during one episode can be reasoned about later.

We explore both a discrete VLN setting based on Room-to-Room~\cite{anderson:cvpr18} episodes and navigation graphs (\benchmark) and a continuous simulation VLN-CE~\cite{krantz_vlnce_2020} setting (\benchmarkce).
The markedly different action and visual observation spaces of these settings may require different memory mechanisms.
In the discrete setting, agents move on graph edges and observe clear, well-framed images.
For \benchmark, we extend a state-of-the-art transformer agent~\cite{chen2021history} that learns an implicit memory based on path history when interpreting instructions.
In the continuous setting, agents take motion actions while observing noisy images of a 3D environment reconstructed from discrete panorama images.
For \benchmarkce, we propose an agent that builds and interprets an explicit semantic map.

In short, we define \paradigmfull\ (\paradigm), a paradigm for persistent VLN, and release \benchmark\ and \benchmarkce\ to study discrete and continuous navigation agents in the \paradigm\ setting.
We create initial agents for both benchmarks, including explicit mapping and implicit memory models for continuous navigation.
Please see \href{https://jacobkrantz.github.io/ivln}{jacobkrantz.github.io/ivln} for code and more details.

\begin{figure*}[t]
    \centering
    \includegraphics[width=0.95\linewidth]{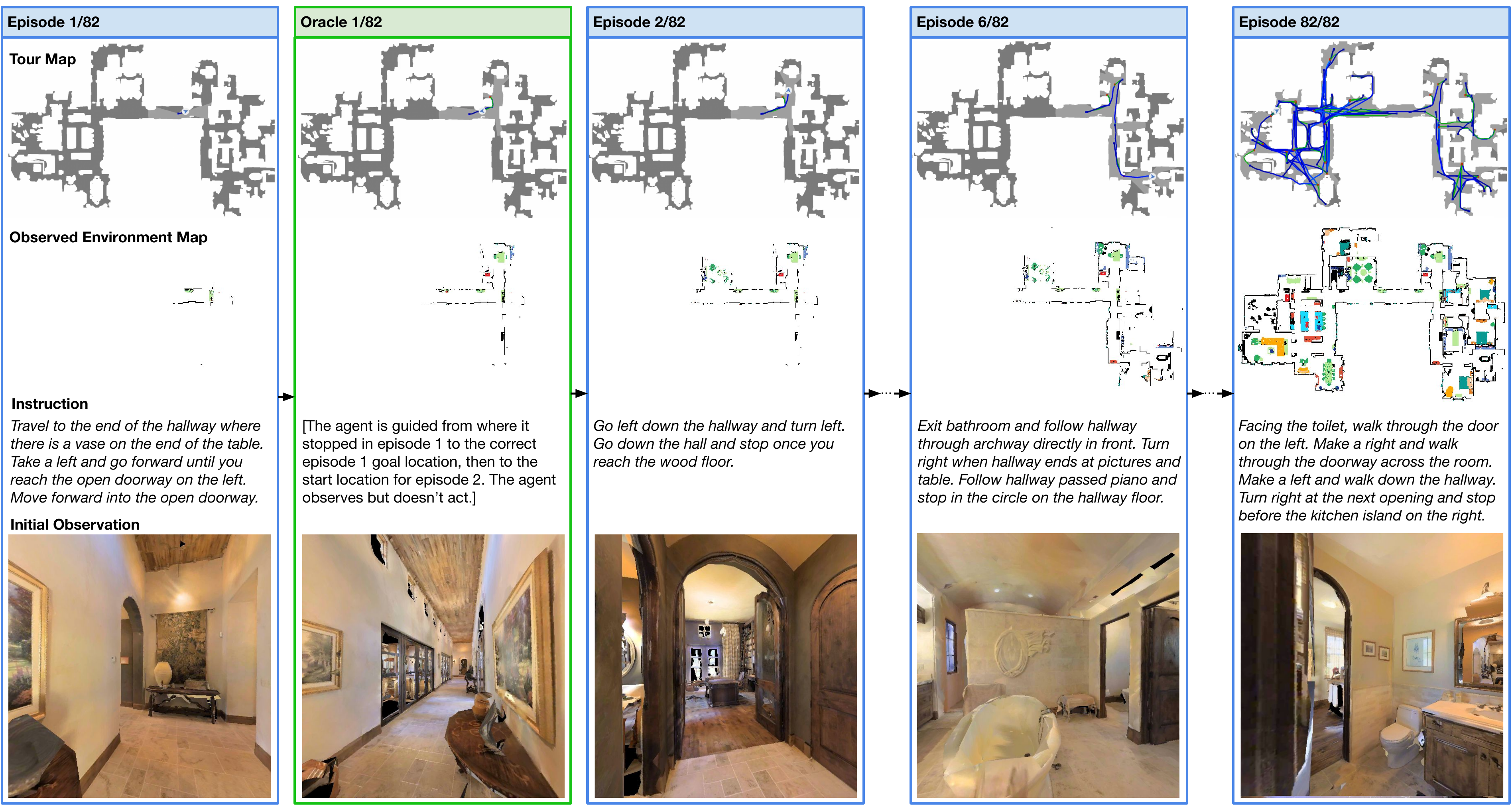}
    \caption{
        In \paradigm, agents are given language instructions corresponding to a sequence of paths that form a \textit{tour} around a 3D scene.
        After attempting to follow each instruction, the agent is teleoperated by an oracle to the correct goal location, then to the start of the next path where the next instruction is issued. Unlike conventional \textit{episodic} paradigms, the agent retains memory between episodes.
    }
    \label{fig:itervlnce}
\end{figure*}

\section{Related Work}
\label{sec:related}

Instruction-guided navigation is a growing area in grounded language understanding with many task settings developed \cite{chen:cvpr19,anderson:cvpr18,shridhar:cvpr20,cvdn,rxr,qi2020reverie}. 
Among these, the Vision-and-Language Navigation (VLN) task setting based on the Room-to-Room (R2R) dataset \cite{anderson:cvpr18} has become a popular benchmark. 
An agent in VLN must follow a natural language instruction by navigating along the described path in a \emph{never-before-seen environment}.
By design, this paradigm does not consider how persistent agents operating over time might leverage prior experiences to better follow future instructions within the same environment. In contrast, accumulating prior experience within an environment is a staple of robotic deployment -- e.g.~building semantic maps for localization and reasoning \cite{rosinol2020kimera, tellex:arcras:20}. Our \paradigm\ paradigm is designed to better align VLN with a realistic robotic deployment scenario.

\paragraph{Benchmarks for VLN in Discrete Settings} VLN tasks frequently involve inferring agent actions in a rendered 2D or 3D scene in response to language commands~\cite{macmahon:aaai06,chen:aaai11}.
Agent control is typically limited to changing position and orientation by discrete amounts or to predefined possible options.
Advances in camera technology have enabled language-guided navigation in photorealistic indoor scenes~\cite{anderson:cvpr18,chang2017matterport3d} and outdoor city spaces~\cite{chen:cvpr19}.
In ``Room-to-Room'' (R2R)~\cite{anderson:cvpr18} VLN, an agent interprets a single English instruction to navigate along a short, indoor path.
In a survey of VLN modeling methods, environment exploration and memorization were identified as frequent strategies for aligning a language instruction to a desired goal location in a scene~\cite{gu:acl22}.
However, R2R evaluates policies on single instructions, limiting the incentive to perform efficient, effective memorization or mapping.
To study longer horizon planning, researchers have extended R2R by concatenating language-aligned paths and their associated instructions~\cite{jain-etal-2019-stay,babywalk}, tasking agents not just with arriving to the goal but with following closely the described path.
Others have collected longer paths with instructions in three languages~\cite{rxr} or given as a cooperative conversation~\cite{cvdn}.
With \benchmark\ \textit{tours}, we present the longest such paths with substantial overlap in areas- covered-before through time, challenging researchers to  utilize information from prior instructions and experience in the scene.

\paragraph{Benchmarks for VLN in Continuous Settings} 
Moving a physical robot, such as a quad-copter~\cite{blukis:corl18} or a toy car~\cite{banerjee2020robotslang}, in response to language instructions requires contending with the real, continuous world.
Existing work has transferred policies for discrete VLN to the physical world by manually curating a discrete representation of the world map as a navigation graph~\cite{anderson2020sim} with limited success.
VLN-CE~\cite{krantz_vlnce_2020} re-introduces Room-to-Room~\cite{anderson:cvpr18} with a \textit{continuous}, 3D reconstruction of indoor MatterPort3D scenes.
However, VLN-CE evaluates agents on single instructions and associated paths in an \iid\ fashion. In contrast, our \benchmarkce\ benchmark incentivizes policies that respect environment persistence found in the real world.
Beyond removing the abstractions of discrete VLN (VLN-CE), \benchmarkce\ situates agents in a scene for long time horizons with many language instructions; a logical next step towards learning useful world representations through visual and linguistic information.

\paragraph{Pre-Exploration in VLN} Some approaches in VLN have embraced a setting where agents can fully explore the environment before following an instruction, either explicitly through pretraining (e.g.~\cite{zhu2020vision, tan2019learning, wang2019reinforced}) or through beam-search at inference time (e.g.~\cite{majumdar2020improving,fried2018speaker}). 
Pre-exploration methods outperform standard VLN approaches and serve as a natural upper bound to \paradigm\ where an agent has fully explored the environment. 
In contrast, \paradigm\ studies how environment information can be collected while performing the task (rather than a priori) and how this partial, opportunistic information can be leveraged to perform better over time.

\paragraph{Persistent Environments in Embodied AI}
Zooming out, visual navigation tasks in embodied AI have seen significant progress, fueled by increased scale and quality of 3D scene datasets (e.g.~\cite{chang2017matterport3d,ramakrishnan2021habitat}) and high-performance simulation platforms (e.g.~\cite{ai2thor, savva2019habitat, xiazamirhe2018gibsonenv, makoviychuk2021isaac}). A focus on real-world complexity has emerged. One recognition is that agents act in, and interact with, persistent environments. Tasks such as multi-object navigation~\cite{wani2020multion} and visual room rearrangement~\cite{weihs2021visual} involve solving sequences of subtasks that, when approached independently, cannot be solved optimally. Instead, reasoning over persistent semantic and spatial information is required. The proposed \paradigm\ paradigm enriches this scene perception problem with natural language and enables the association of persistent visual semantics with linguistic information.

\section{\paradigmfull{}}
\label{sec:paradigm}

We facilitate the study of agents given sequential navigation instructions in natural language.
We extend the Room-to-Room (R2R)~\cite{anderson:cvpr18} dataset of independent \textit{episodes}---natural language instructions and associated target paths in a particular scene---to \textit{tours}---sequences of many episodes that cover large swaths of the scene and include backtracking.
The resulting \benchmarkfull\ tours contain substantially longer paths and navigation instruction context than prior discrete (\benchmark) or continuous (\benchmarkce) VLN benchmarks.

\paragraph{The Iterative Paradigm}
We define a \textit{tour} to be an ordered sequence of episodes within a scene.
Tours alternate between two phases.
In the \textit{agent navigation} phase, the agent is given a language instruction and infers navigation actions, equivalent to a VLN \textit{episode}. 
The phase ends when the agent emits the STOP signal or takes a maximum number of actions.
The \textit{oracle navigation} phase immediately follows in two parts.
First, if the agent has not successfully navigated to within 0.5m of the episode goal, it is guided without language to that goal by an oracle that forces its actions, analogous to a human teaching the robot where the path should have ended.
Second, the agent is oracle-guided to the starting point of the next episode in the tour, analogous to following a human and waiting to receive the next instruction.
The agent passively observes the environment during this phase.

\paragraph{Generating Tours from VLN Data}
We generate tours that minimize the distance between end and start points of sequential episodes. 
We also maximize the number of included episodes as path finding between poses can fail in \benchmarkce.

Each R2R split contains a set of scenes, which each contain a set of episodes $E$. 
For each $E$, we seek to derive a set of disjoint tours $\mathcal{T}$ where each tour $T \in \mathcal{T}$ is a sequence of episodes that can be inter-navigated. 
That is, for episode $i$ and $i+1$ in $T$, navigation from the end of $i$ to the start of $i+1$ is possible. Letting $X$ be the set of unique paths in an episode set $E$, we first partition $P(X)$ such that the paths in each subset $p$ are inter-navigable; closed doors or obstacles can create disjoint regions in the scene. To determine $P(X)$, we compute the navigable geodesic distance between each path pair where a finite distance implies connectivity. In \benchmark, this distance is computed on a navigation graph; in \benchmarkce, it is computed on a 3D navigation mesh and assumes agent dimensions and actions common to VLN-CE \cite{krantz_vlnce_2020}. We then order the paths in each subset $p$ to define a tour $T$. Minimizing the oracle navigation distance in a tour is equivalent to an asymmetric traveling salesperson problem (ATSP) which we approximately solve using the Lin-Kernighan heuristic (LKH)~\cite{helsgaun2000effective}. Finally, if $E$ contains $n$ instructions per path and $n>1$, we duplicate each tour $n$ times, sampling an instruction for each path without replacement. 

\begin{figure*}[t]
    \centering
	\resizebox{0.85\textwidth}{!}{
    \begin{subfigure}{0.3\textwidth}
        \includegraphics[width=\textwidth]{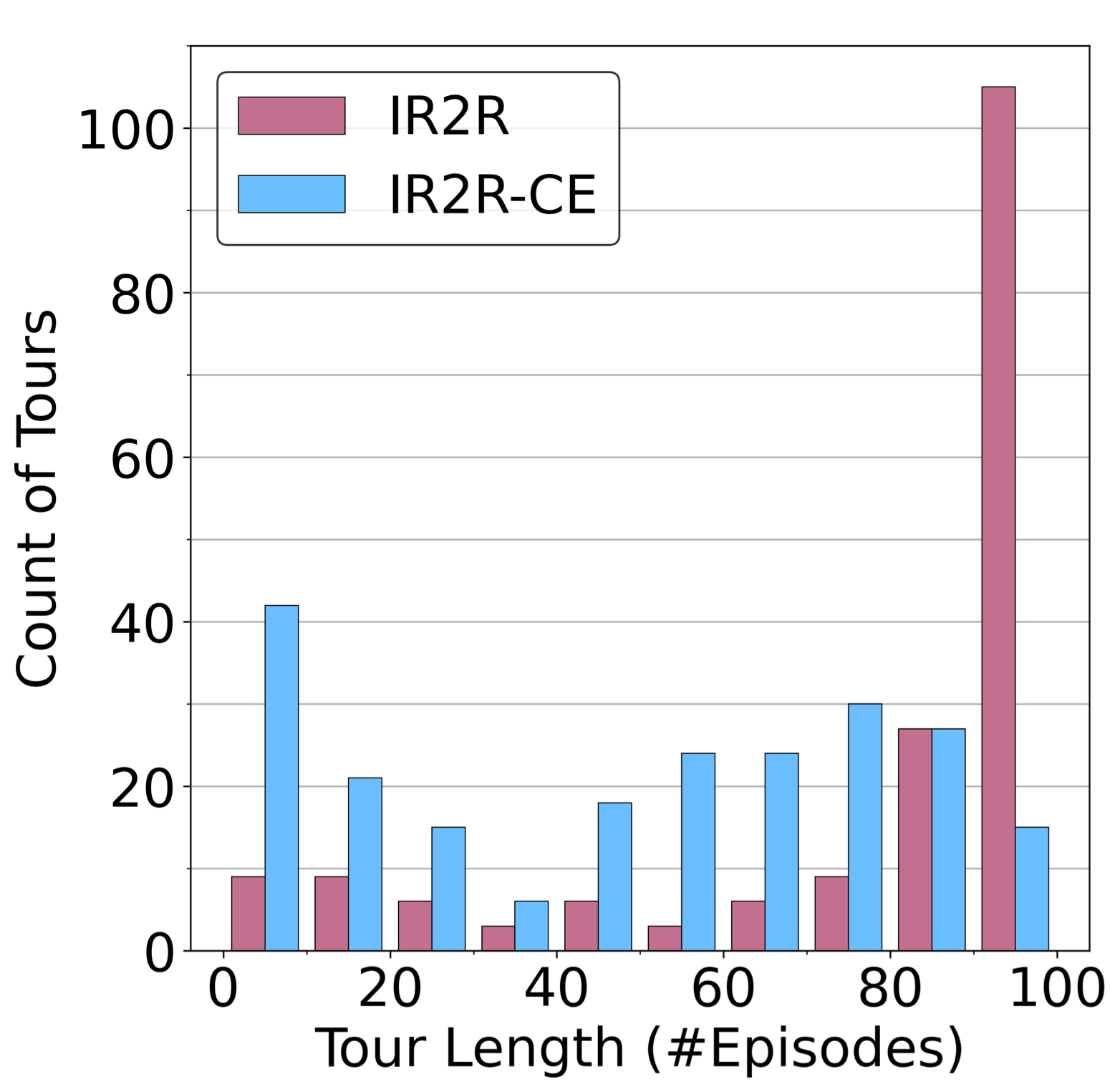}
        \caption{Episodes per tour.}
        \label{fig:tour_lengths}
    \end{subfigure}\hspace{0.8in}
    \begin{subfigure}{0.55\textwidth}
        \includegraphics[width=\textwidth]{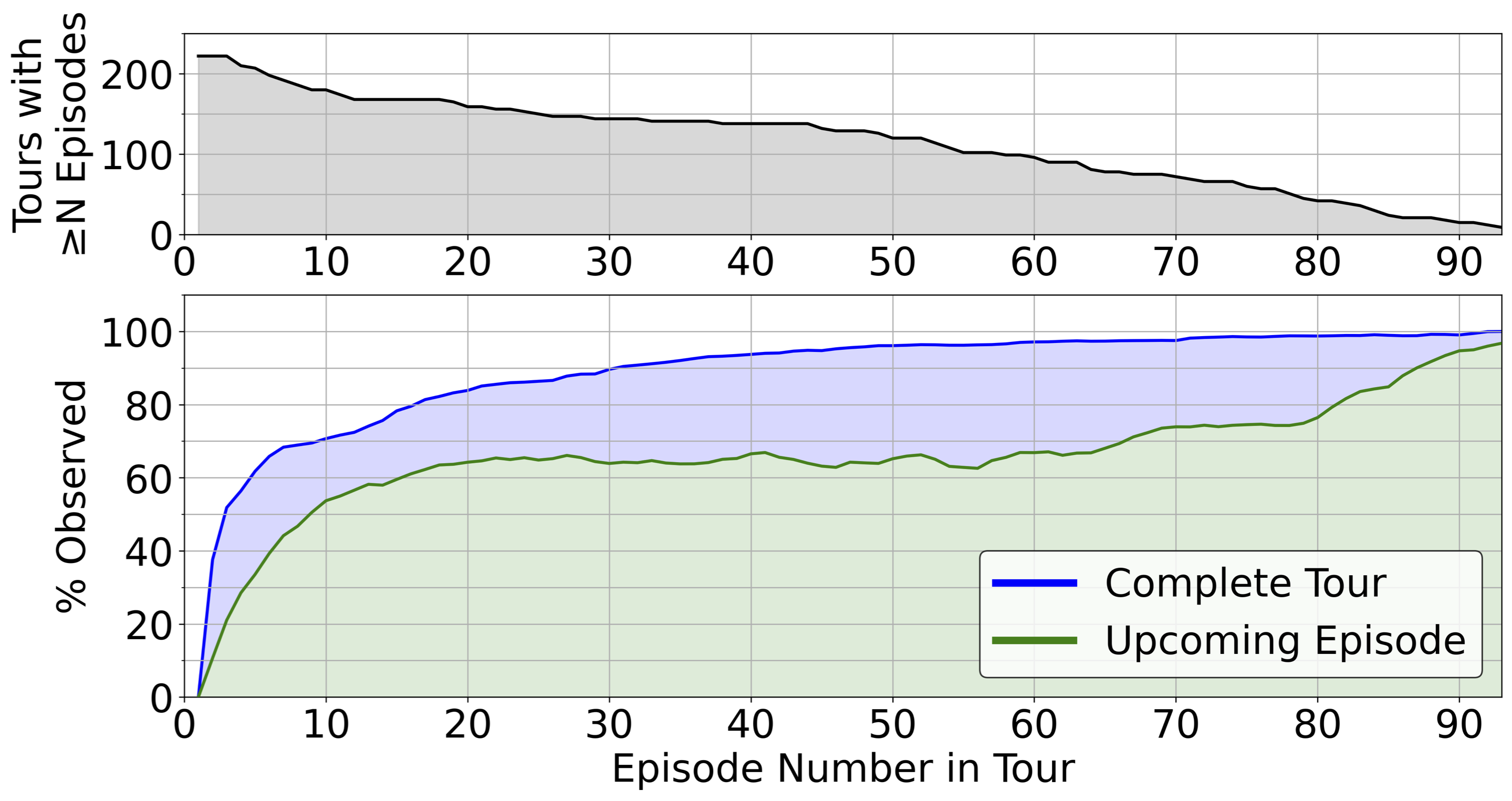}
        \caption{\benchmarkce\ observation coverage relative to tour progress.}
        \label{fig:tour_obs}
    \end{subfigure}\hfill
    }
    \caption{
        (a) We compare the distributions of tour lengths between the \benchmark{} and \benchmarkce{} Train splits.
        (b) We consider an oracle agent following the target paths of each tour in \benchmarkce{} Train. 
        Before starting each episode, we measure the percentage of that episode's target path observed earlier in the tour (\textcolor{pltgreen}{Upcoming Episode}). 
        To do this, we compute what percentage of an episodic coverage map is accounted for in a map iteratively constructed by the oracle agent.
        We also measure what percentage of the entire tour has been observed (\textcolor{blue}{Complete Tour}).
    }
    \label{fig:tour_dist_obs}
\end{figure*}

\paragraph{Dataset Characteristics}
\setlength{\tabcolsep}{.3em}
\begin{table}[t]
	\begin{center}
	\resizebox{\columnwidth}{!}{
	    \small
		\begin{tabular}{llrr rrrrrr}
			\toprule
            \multirow{2}{*}{ \shortstack{Dataset}}
            & \multirow{2}{*}{ \shortstack{Split}}
            & \multirow{2}{*}{ \shortstack{Scenes}}
            & \multirow{2}{*}{ \shortstack{Episodes}}
            & \multirow{2}{*}{ \shortstack{Tours}}
            & \multirow{2}{*}{ \shortstack{Tours/\\Scene}}
			& \multicolumn{4}{c}{\textbf{Tour Length (Episodes)}}
            \\
			\cmidrule{7-10}
			&&&&&& \textbf{\texttt{Mean}}
			& \textbf{\texttt{Min}}
			& \textbf{\texttt{Max}}
			& \textbf{\texttt{SD}}
			\\
			\midrule
			\multirow{3}{*}{ \shortstack{\benchmark}}
            & { Train} & 61 & 14025 & 183 & 3.0 & 76.6 & 2 & 99 & 28.4
			\\
		    & { Val-Seen} & 53 & 1011 & 159 & 3.0 & 6.4 & 2 & 11 & 2.1
			\\
		    & { Val-Unseen} & 11 & 2349 & 33 & 3.0 & 71.2 & 6 & 100 & 34.0
			\\
			\midrule
			\multirow{3}{*}{ \shortstack{\benchmarkce}}
            & { Train} & 60 & 10668 & 222 & 3.7 & 48.1 & 3 & 93 & 30.5
			\\
		    & { Val-Seen} & 50 & 747 & 156 & 3.1 & 4.8 & 2 & 10 & 2.1
			\\
		    & { Val-Unseen} & 11 & 1824 & 36 & 3.3 & 50.7 & 3 & 100 & 31.3
			\\
			\bottomrule
		\end{tabular}
	}
	\end{center}
	\caption{We construct sequences of episodes---\textit{tours}---from the Room-to-Room dataset \cite{anderson:cvpr18} to create the discrete \benchmark\ and continuous \benchmarkce\ benchmarks. Here we detail characteristics of these benchmarks, including the average number of episodes per  tour.}
	\label{tab:tour_stats}
\end{table}
We generate tours in the Train, Validation-Seen, and Validation-Unseen splits of discrete R2R to form \benchmark\ and continuous R2R to form \benchmarkce\ (\tabref{tab:tour_stats}).
Validation-Seen (Val-Seen) contains episodes from scenes seen during training, while Validation-Unseen (Val-Unseen) contains episodes from scenes not seen during training.
In total, \benchmark{} contains 375 tours and \benchmarkce{} contains 414. 
There are fewer discrete tours, which are longer on average than continuous tours (\figref{fig:tour_lengths}),
due to discontinuities in the navigable area of continuous environments.
In discrete VLN, a path exists from each node to every other node in a scene, but in continuous environments navigation between episode endpoints can fail, resulting in disjoint spaces within a scene that have shorter tours.
The distribution of episodes per tour has a high variance for both benchmarks, a reflection of path sampling in R2R and a diversity of scene sizes.

Since episodes are chained together in \paradigm, locations along the target path for the current episode may have been observed earlier in the tour.
In \figref{fig:tour_obs}, we visualize how much of a tour's scene region has been observed relative to the number of episodes performed. 
We find that following target paths quickly observes a majority of the region; after just 10 episodes, on average, over 50\% of the next target path and over 70\% of the entire tour region have been observed.

\begin{figure*}[t]
    \centering
    \begin{subfigure}{0.31\textwidth}
        \includegraphics[width=\textwidth]{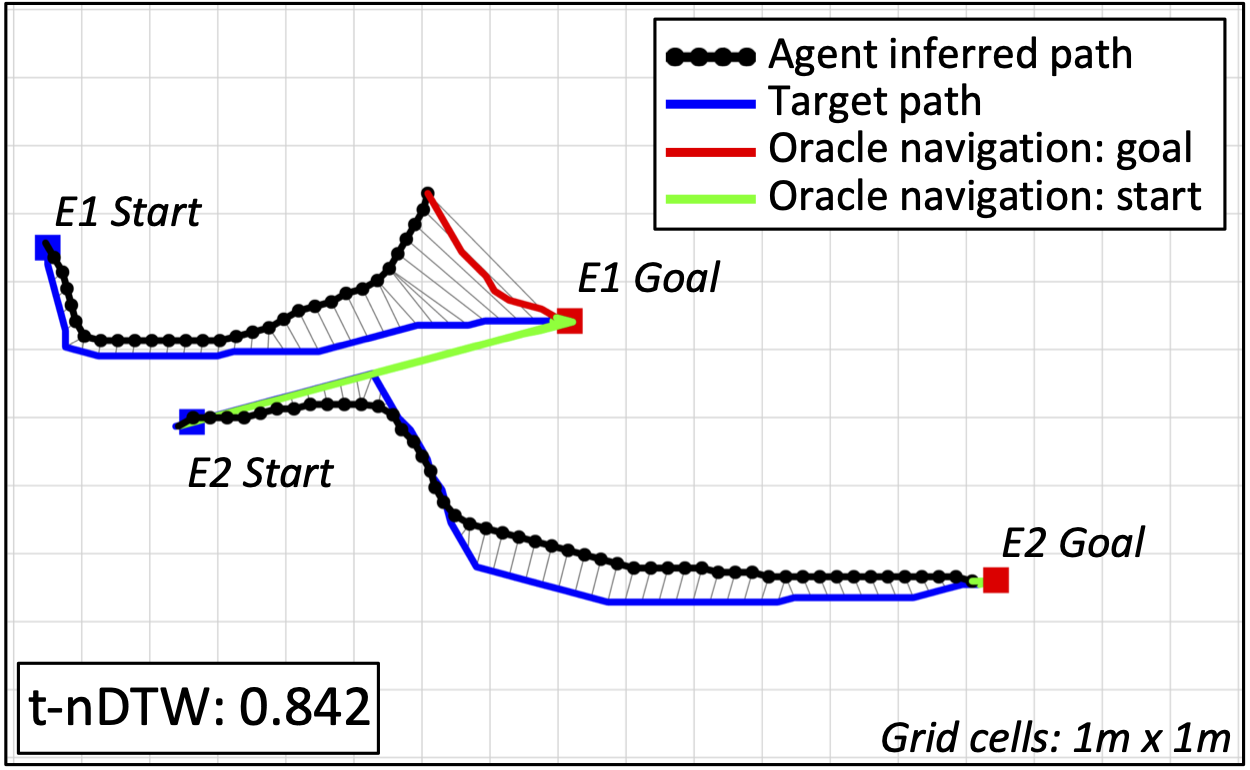}
        \caption{}
        \label{fig:ndtw1}
    \end{subfigure}\hspace{0.25em}
    \begin{subfigure}{0.31\textwidth}
        \includegraphics[width=\textwidth]{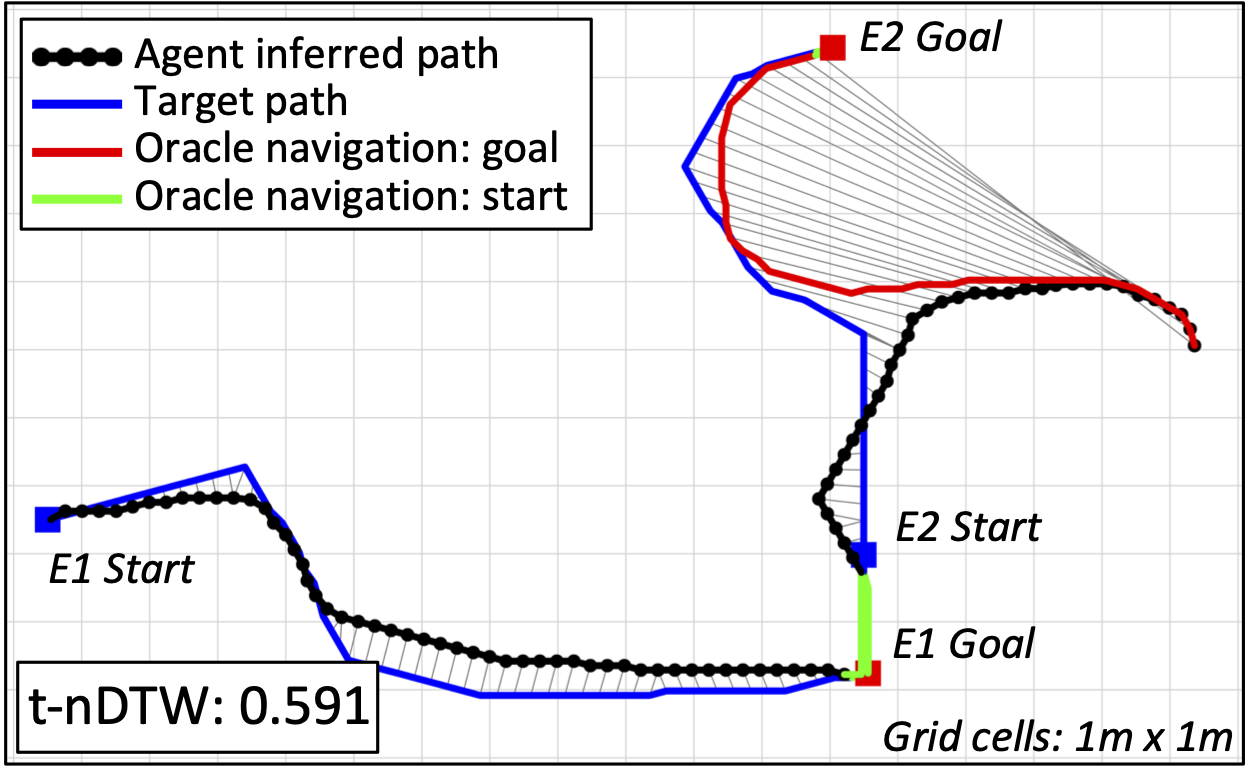}
        \caption{}
        \label{fig:ndtw2}
    \end{subfigure}\hspace{0.25em}
    \begin{subfigure}{0.31\textwidth}
        \includegraphics[width=\textwidth]{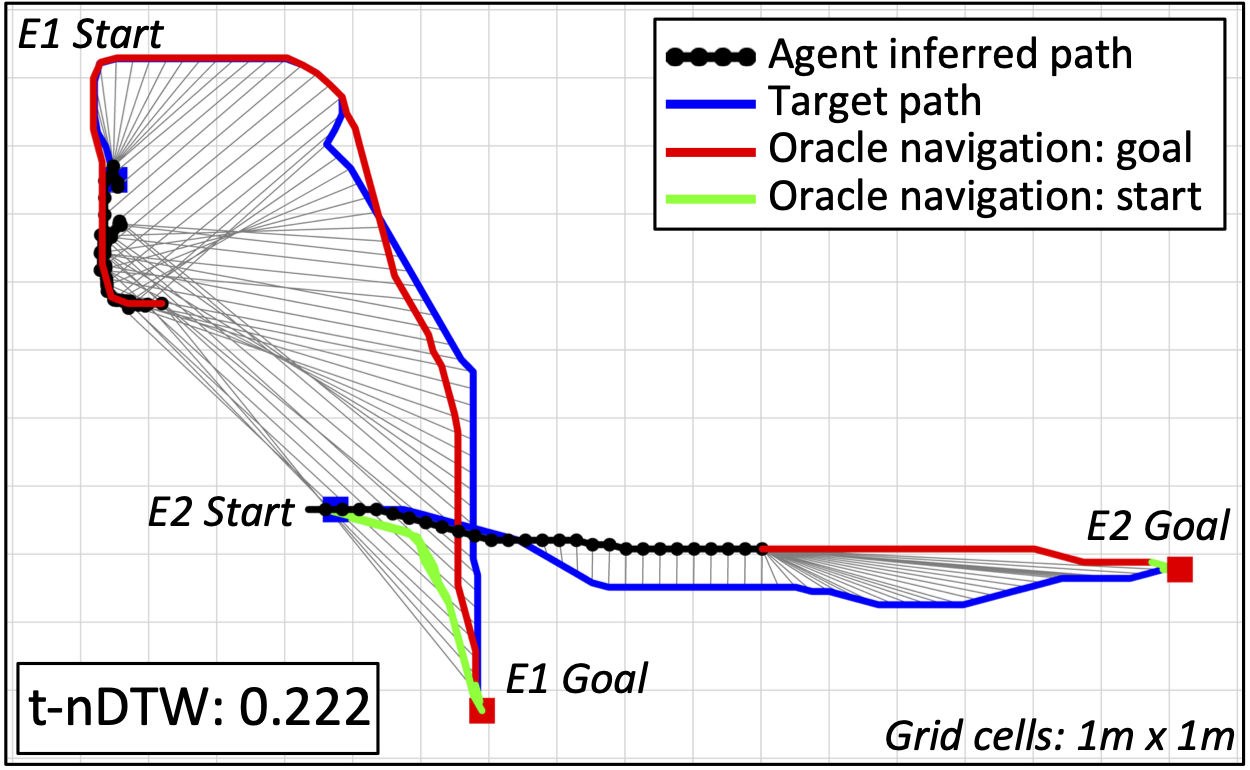}
        \caption{}
        \label{fig:ndtw3}
    \end{subfigure}
    \caption{
        Evaluations in \benchmarkce\ using t-nDTW for example 2-episode tours.
        We visualize the DTW alignment of the \textbf{agent's inferred path} by its match to the \textcolor{blue}{target path}.
        Between episodes, an oracle conveys the agent \textcolor{red}{from inferred episode stop point to true stop}, \textcolor{pltgreen}{then on to the next episode start}.
        In \figref{fig:ndtw3}, poor performance in episode 1 (E1) drops the overall t-nDTW score significantly.
    }
    \label{fig:ndtw}
\end{figure*}

\paragraph{Metrics for Iterative Evaluation}
Agents should be successful and efficient from the start of a tour and improve as the tour progresses. 
For iterative (tour-based) evaluation, we adapt the normalized dynamic time warping (nDTW \cite{magalhaes2019effective}) metric. We select nDTW over success-based metrics for unification across \paradigm\ benchmarks: \benchmark\ and \benchmarkce\ contain shortest-path priors (all instructions describe the shortest path to the goal), but other datasets that can be converted to \paradigm\ do not, such as RxR \cite{rxr}.
In episodic VLN, the nDTW metric is computed per-episode and averaged across all episodes. 
However, in \paradigm, episodes within a tour need collective evaluation.
Otherwise, an agent could exploit per-episode averaging by using the first episode to explore the entire environment, improving subsequent performance at the cost of poor performance in just a single episode---when the number of episodes in a scene is large, this would have minimal impact on averaged metrics.

Formally, given a candidate path $Q$ and a target path $R$, each consisting of a sequence of 3D points, nDTW computes the dynamic time warping (DTW) cost between $Q$ and $R$ normalized by the number of points in the reference path ($|R|$) and a distance threshold of success ($d_{th}$):
\begin{align}
    \texttt{nDTW}(R,Q) = \text{exp} \left ( - \frac{\texttt{DTW}(R, Q)}{|R| \cdot d_{th}} \right )
\end{align}
As in tour generation, we use navigation graph distance in \benchmark\ and geodesic distance in \benchmarkce.

To extend this definition to tours, we make two changes to the episodic nDTW calculation. First, we compute nDTW over tour paths $Q^T$ and $R^T$ instead of episode paths $Q$ and $R$. The candidate path $Q^T$ includes the points visited during agent navigation phases of tour $T$, and the target path $R^T$ is the concatenation of the target paths for each episode in $T$. Oracle navigation is excluded from both $Q^T$ and $R^T$. Second, to ensure that episode boundaries are respected when aligning candidate and target points in the DTW calculation, candidate and target points from $Q^T$ and $R^T$ are assigned infinite distance unless they belong to the same episode in the tour. This penalty ensures that an agent can't receive credit for completing a path while following a different instruction.

To compute performance for a dataset split, we aggregate nDTW weighted by episode count in each tour $T_i$, avoiding inflated scores from performing well only on short tours:
\begin{align}
    \texttt{t-nDTW} = \sum_i \frac{|T_i| \cdot \texttt{nDTW}(R^{T_i}, Q^{T_i})}{\sum_j |T_j|}
\end{align}
The tour nDTW score (\metric{t-nDTW}) is bounded between 0 and 1, with 1 indicating perfect alignment of the agent's path and the target path for every episode of every tour in the split. 
In the following experiments, we report \metric{t-nDTW} scaled between 0 and 100 as is common practice for episodic \metric{nDTW}.
\metric{t-nDTW} functions in discrete and continuous environments and serves as the primary metric in \benchmark{} and \benchmarkce{}.
\figref{fig:ndtw} contains example \metric{t-nDTW} evaluations of an \benchmarkce{} agent to illustrate the relationship between \metric{t-nDTW} and path alignment. We include more examples in the appendix.

\section{Methods}
\label{sec:methods}

We demonstrate how VLN and VLN-CE baseline models generalize to our iterative task
and explore whether adding persistent \textit{tour memory} (either unstructured latent memory or a spatial semantic map) improves performance. 

\subsection{VLN Baseline Agents}
\label{sec:vln_methods}


\paragraph{HAMT} 
We adopt the History-Aware Multimodal Transformer (HAMT)~\cite{chen2021history} agent. Like many recent methods, HAMT is a transformer-based agent pretrained on proxy tasks and finetuned on VLN.
HAMT has three transformer-based encoders: an instruction encoder, a visual encoder for the current observation, and a history encoder for previous state-action pairs.
A cross-modal transformer fuses these to predict the next action.
The history embedding at time step $t$ is represented as $\{ h_\texttt{CLS}, h_{1}, ..., h_{t-1} \}$, where $h_t$ is the features of the state-action pair at step $t$ and $h_\texttt{CLS}$ is the features of the $\texttt{CLS}$ token used to gather sequence-level information.

\paragraph{\thamt}
We enable tour-level reasoning by including state-action pairs from previous episodes in the history embedding.
For episode $i$, we denote the total number of steps as $l_i$, including the oracle navigation after termination.
We denote the state-action embedding at step $t$ for episode $i$ as $h^{i}_{t}$. 
At step $t$ in $i$, we set the history embedding as $\{ h_\texttt{PREV}, h^{1}_{1}, ..., h^{1}_{l_1}, ..., h^{i-1}_{1}, ..., h^{i-1}_{l_{i-1}}, h_\texttt{CLS}, h^{i}_{1}, ..., h^{i}_{t-1} \}$, where \texttt{PREV} is a token delineating episode boundaries. We limit to the latest 50 steps.
Unlike the original HAMT, we unfreeze the history encoder to learn this modified history encoding. 
We train via teacher-forcing with inflection weighting~\cite{wijmans2019emboied} and update gradients per episode in a tour.

\subsection{VLN-CE Baseline Agents}
\label{sec:vlnce_methods}

The Cross-Modal Attention (CMA) agent defined in VLN-CE \cite{krantz_vlnce_2020} is a common baseline in recent works \cite{irshad2021hierarchical, irshad2022sasra, krantz2021waypoint, hong2022bridging, georgakis2022cm2, chen2021topological}.
CMA is an end-to-end recurrent model that observes RGBD, the instruction, and the previous action to predict an action from \action{TURN-LEFT}, \action{TURN-RIGHT}, \action{MOVE-FORWARD}, and \action{STOP}. CMA has a two-GRU structure to track episodic history; one tracks vision and the other tracks general state from which the action is predicted.

\subsubsection{Agents with Unstructured Latent Memory}

\begin{figure*}[t]
    \centering
    \includegraphics[width=.85\textwidth]{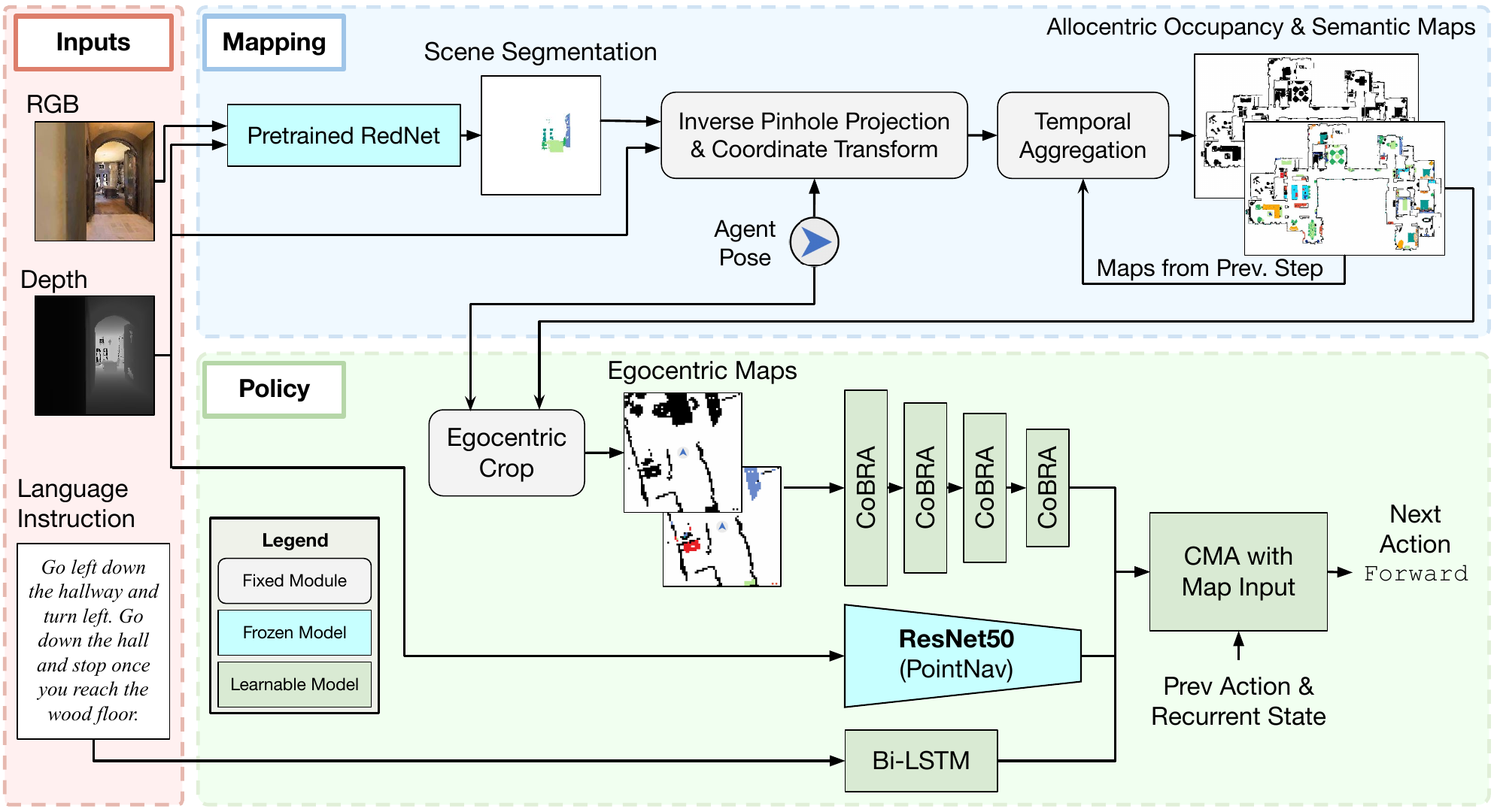}
    \caption{
        In addition to the encoders for language instructions and depth frames, MAP-CMA model learns an encoding of an egocentric crop of a top-down semantic map of the environment constructed by the agent during navigation in order to predict the next navigation action.
    }
    \label{fig:mapcma}
\end{figure*}

\paragraph{CMA}
We consider adaptations of the CMA agent's unstructured latent memory to tours. Against these, the original CMA agent is included as a baseline with no cross-episode reasoning ability; that is, hidden states are reset each episode.

\paragraph{TourCMA}
We reset the hidden state of the vision GRU only at the start of each tour, thereby extending the temporal receptive field to all tour steps. We reset the state GRU episodically.
This model provides structure for reasoning over both scene-level vision and episode-specific visuo-linguistics.

\paragraph{PoolCMA}
We enable both tour and episodic memory within the vision GRU. We reset the hidden state each episode but temporally max-pool the hidden state into a tour-persistent vector reset each tour. 
This vector is input to the vision GRU. 

\paragraph{PoolEndCMA}
Observations from previous episodes provide utility if they are relevant for planning, \eg, when considering a previously-traversed hallway.
Agents may learn to ignore this signal in favor of episode-specific alignments.
We encourage this signal by coupling tour memory to action prediction; we concatenate the \model{PoolCMA} tour memory with the final state vector and use the result to predict an action. 

\paragraph{Training Method}
We train the above models in \benchmarkce\ using teacher forcing and update parameters following each episode rollout.
For tour-persistent memory structures, we disable gradients from steps prior to the current episode. 
This equates to an adaptive truncated backpropagation through time (TBPTT) where the time step is episode length.

\subsubsection{MAP-CMA: Agents with Semantic Maps}

We experiment with providing agent-centric, metrically accurate local crops of maps of an agent's surroundings to our models to evaluate the impact of structured memory. 

\textbf{MAP-CMA} We build occupancy maps \cite{elfes1989using} where cells are either 0 (empty) or 1 (occupied), and semantic maps contain one-hot vectors of thirteen common labels in R2R environments~\cite{cartillier2020semantic}.
Following~\cite{cartillier2020semantic}, we use the inverse pinhole camera projection model to unproject depth measurements to 3D pointclouds. We also unproject egocentric semantics---both ground-truth \cite{chang2017matterport3d} and from a fine-tuned Rednet feature encoder \cite{cartillier2020semantic, jiang2018rednet}---to form a semantic pointcloud.
We collapse these pointclouds along the height dimension to generate 2D maps.
When more than one semantic label exists in a height column, we choose the highest semantic label available to project. 
Agents may traverse between floors. As such, we constrain pointclouds to project only features lying between the floor and ceiling planes relative to the agent’s 3D pose.

We augment the existing CMA architecture with semantic and occupancy maps by replacing the RGB input with maps. 
Specifically, we channel-wise concatenate the semantic and occupancy maps, encode them through a learned convolutional encoder, and produce a spatial embedding that propagates through CMA in place of RGB features.
The map input is $14{\times}64{\times}64$, representing a $64{\times}64$ spatial grid with 13 one-hot semantic channels and one occupancy channel.
Four convolutional blocks each consisting of a \textbf{Co}nvolution, \textbf{B}atch normalization, \textbf{R}eLU activation, and \textbf{A}verage pool (\textbf{CoBRA} in \figref{fig:mapcma}) encode the map to a $128{\times}4{\times}4$ output.
These semantic spatial features are used as a drop-in replacement of visual features in CMA as depicted in \figref{fig:mapcma}.

\paragraph{Training Method}
We train this model in \benchmarkce\ following the two-step method presented in \cite{krantz_vlnce_2020} used to train the CMA model. We initially train using teacher forcing on the augmented EnvDrop \cite{tan2019learning} ported to \benchmarkce, then fine-tune the best-performing EnvDrop Val-Unseen checkpoint using DAgger \cite{ross2011reduction} on \benchmarkce\ Train. The best performing checkpoint on Val-Unseen is accepted as the final model. We use the Progress Monitor \cite{ma2019self} auxiliary loss in both training steps. 
We train with episodic maps (reset each episode), iterative maps (reset each tour), and known maps (pre-computed for each scene). 
For each temporal construction method, we train one model on ground-truth semantic labels and one model with predicted semantics, resulting in 6 total trained models capable of evaluating the impacts of temporal map construction and semantic segmentation in \benchmarkce.

\section{Experiments and Results}
\label{sec:results}

\setlength{\tabcolsep}{.3em}
\begin{table*}[t]
    \renewcommand{\arraystretch}{1.15}
    \setlength{\aboverulesep}{0pt}
    \setlength{\belowrulesep}{0pt}
    \begin{center}
	\resizebox{0.92\textwidth}{!}{
		\begin{tabular}{clcccc c ccccccs c ccccccs}
			\toprule
            & & & & & &
			& \multicolumn{7}{c}{\scriptsize\textbf{Val-Seen}}
		   && \multicolumn{7}{c}{\scriptsize\textbf{Val-Unseen}}
            \\
			\cmidrule{8-14}
			\cmidrule{16-22}
			\scriptsize \shortstack{\#} &
			{\scriptsize Model}
			& \footnotesize \textsc{ph}
			& \footnotesize \textsc{th}
			& \footnotesize \textsc{phi}
			& \footnotesize \textsc{iw}
			&
			& \scriptsize\textbf{\texttt{TL}}
			& \scriptsize\textbf{\texttt{NE}}~$\downarrow$
			& \scriptsize\textbf{\texttt{OS}}~$\uparrow$
			& \scriptsize\textbf{\texttt{nDTW}}~$\uparrow$
			& \scriptsize\textbf{\texttt{SR}}~$\uparrow$
			& \scriptsize\textbf{\texttt{SPL}}~$\uparrow$
			& \scriptsize\textbf{\texttt{t-nDTW}}~$\uparrow$
			&
			& \scriptsize\textbf{\texttt{TL}}
			& \scriptsize\textbf{\texttt{NE}}~$\downarrow$
			& \scriptsize\textbf{\texttt{OS}}~$\uparrow$
			& \scriptsize\textbf{\texttt{nDTW}}~$\uparrow$
			& \scriptsize\textbf{\texttt{SR}}~$\uparrow$
			& \scriptsize\textbf{\texttt{SPL}}~$\uparrow$
			& \scriptsize\textbf{\texttt{t-nDTW}}~$\uparrow$
			\\
			\midrule
			\scriptsize \texttt{1}
                & \texttt{HAMT}
                & & & & &
                & 10.1 \scriptsize{$\pm$0.1}
                & \bf 4.2 \scriptsize{$\pm$0.1}
                & \bf 70 \scriptsize{$\pm$1}
                & \bf 71 \scriptsize{$\pm$1}
                & \bf 63 \scriptsize{$\pm$1}
                & \bf 61 \scriptsize{$\pm$1}
                & \bf 58 \scriptsize{$\pm$1}
                &
                & \hphantom{0}9.4 \scriptsize{$\pm$0.1}
                & \bf 4.7 \scriptsize{$\pm$0.0}
                & \bf 64 \scriptsize{$\pm$1}
                & \bf 66 \scriptsize{$\pm$0}
                & \bf 56 \scriptsize{$\pm$0}
                & \bf 54 \scriptsize{$\pm$0}
                & \bf 50 \scriptsize{$\pm$0}
			\\
			\scriptsize \texttt{2}
			    & \texttt{TourHAMT}
			    & \checkmark & \checkmark & \checkmark & \checkmark &
			    & \hphantom{0}9.4 \scriptsize{$\pm$0.4}
                & 5.8 \scriptsize{$\pm$0.1}
                & 56 \scriptsize{$\pm$1}
                & 59 \scriptsize{$\pm$0}
                & 45 \scriptsize{$\pm$1}
                & 43 \scriptsize{$\pm$1}
                & 45 \scriptsize{$\pm$0}
                &
                & 10.0 \scriptsize{$\pm$0.2}
                & 6.2 \scriptsize{$\pm$0.1}
                & 52 \scriptsize{$\pm$2}
                & 52 \scriptsize{$\pm$0}
                & 39 \scriptsize{$\pm$1}
                & 36 \scriptsize{$\pm$0}
                & 32 \scriptsize{$\pm$1}
			\\
			\scriptsize \texttt{3}
			    & 
			    & \checkmark & \checkmark & \checkmark &  &
                & 10.5 \scriptsize{$\pm$0.3}
                & 6.0 \scriptsize{$\pm$0.2}
                & 60 \scriptsize{$\pm$1}
                & 58 \scriptsize{$\pm$1}
                & 45 \scriptsize{$\pm$2}
                & 43 \scriptsize{$\pm$2}
                & 42 \scriptsize{$\pm$1}
                &
                & 10.9 \scriptsize{$\pm$0.2}
                & 6.8 \scriptsize{$\pm$0.2}
                & 54 \scriptsize{$\pm$1}
                & 51 \scriptsize{$\pm$1}
                & 38 \scriptsize{$\pm$1}
                & 34 \scriptsize{$\pm$1}
                & 31 \scriptsize{$\pm$1}
			\\
			\scriptsize \texttt{4}
			    & 
			    & \checkmark & \checkmark &  &  &
                & 10.6 \scriptsize{$\pm$0.3}
                & 6.0 \scriptsize{$\pm$0.1}
                & 61 \scriptsize{$\pm$1}
                & 58 \scriptsize{$\pm$1}
                & 45 \scriptsize{$\pm$1}
                & 42 \scriptsize{$\pm$1}
                & 42 \scriptsize{$\pm$1}
                &
                & 10.3 \scriptsize{$\pm$0.3}
                & 6.7 \scriptsize{$\pm$0.2}
                & 52 \scriptsize{$\pm$1}
                & 50 \scriptsize{$\pm$1}
                & 38 \scriptsize{$\pm$1}
                & 34 \scriptsize{$\pm$1}
                & 29 \scriptsize{$\pm$1}
			\\
			\scriptsize \texttt{5}
			    & 
			    & \checkmark &  &  &  &
                & 10.9 \scriptsize{$\pm$0.3}
                & 6.1 \scriptsize{$\pm$0.1}
                & 60 \scriptsize{$\pm$2}
                & 58 \scriptsize{$\pm$1}
                & 45 \scriptsize{$\pm$1}
                & 42 \scriptsize{$\pm$1}
                & 41 \scriptsize{$\pm$0}
                &
                & 11.0 \scriptsize{$\pm$0.6}
                & 6.7 \scriptsize{$\pm$0.1}
                & 52 \scriptsize{$\pm$2}
                & 51 \scriptsize{$\pm$0}
                & 38 \scriptsize{$\pm$0}
                & 34 \scriptsize{$\pm$0}
                & 28 \scriptsize{$\pm$1}
			\\
			\bottomrule
		\end{tabular}}
	\end{center}
	\caption{
	    \thamt falls short of HAMT on \benchmark.
	    Simple changes to an episodic model are not enough to leverage tours.
	    \textsc{ph}: previous episodes' history; \textsc{th}: trainable history encoder; \textsc{phi}: previous history identifier; \textsc{iw}: inflection weighting. 
	    Metrics are $\bar{x} \pm \sigma_{\bar{x}}$ over 3 runs.
	}
	\label{tab:ivln_results}
\end{table*}
\setlength{\tabcolsep}{.3em}
\begin{table*}[t]
    \setlength{\aboverulesep}{0pt}
    \setlength{\belowrulesep}{0pt}
    \renewcommand{\arraystretch}{1.15}
    \begin{center}
	\resizebox{0.92\textwidth}{!}{
		\begin{tabular}{cl c cccccs c ccccccs}
			\toprule
            & 
			& \multicolumn{7}{c}{\scriptsize\textbf{Val-Seen}}
		   && \multicolumn{7}{c}{\scriptsize\textbf{Val-Unseen}}
            \\
			\cmidrule{3-9}
			\cmidrule{11-17}
			\scriptsize \shortstack{\#} &
			{\scriptsize Model}
			& \scriptsize\textbf{\texttt{TL}}
			& \scriptsize\textbf{\texttt{NE}}~$\downarrow$
			& \scriptsize\textbf{\texttt{OS}}~$\uparrow$
			& \scriptsize\textbf{\texttt{nDTW}}~$\uparrow$
			& \scriptsize\textbf{\texttt{SR}}~$\uparrow$
			& \scriptsize\textbf{\texttt{SPL}}~$\uparrow$
			& \scriptsize\textbf{\texttt{t-nDTW}}~$\uparrow$
			&
			& \scriptsize\textbf{\texttt{TL}}
			& \scriptsize\textbf{\texttt{NE}}~$\downarrow$
			& \scriptsize\textbf{\texttt{OS}}~$\uparrow$
			& \scriptsize\textbf{\texttt{nDTW}}~$\uparrow$
			& \scriptsize\textbf{\texttt{SR}}~$\uparrow$
			& \scriptsize\textbf{\texttt{SPL}}~$\uparrow$
			& \scriptsize\textbf{\texttt{t-nDTW}}~$\uparrow$
			\\
			\midrule
			\scriptsize \texttt{1}
                & \texttt{CMA}
                &  7.8 \scriptsize{$\pm$0.4}
                &  8.8 \scriptsize{$\pm$0.6}
                & 27   \scriptsize{$\pm$3 }
                & 42   \scriptsize{$\pm$3 }
                & 18   \scriptsize{$\pm$3 }
                & 17   \scriptsize{$\pm$3 }
                & 39   \scriptsize{$\pm$1 }
                &&  7.5 \scriptsize{$\pm$0.3}
                 &  8.8 \scriptsize{$\pm$0.2}
                 & \textbf{26   \scriptsize{$\pm$1 }}
                 & \textbf{44   \scriptsize{$\pm$1 }}
                 & \textbf{19   \scriptsize{$\pm$1 }}
                 & \textbf{18   \scriptsize{$\pm$1 }}
                 & \textbf{38   \scriptsize{$\pm$2 }}
			\\
			\scriptsize \texttt{2}
			    & \texttt{TourCMA}
                &  8.0 \scriptsize{$\pm$0.4}
                &  \textbf{8.2 \scriptsize{$\pm$0.9}}
                & \textbf{30   \scriptsize{$\pm$2 }}
                & \textbf{44   \scriptsize{$\pm$2 }}
                & \textbf{20   \scriptsize{$\pm$3 }}
                & \textbf{19   \scriptsize{$\pm$2 }}
                & \textbf{40   \scriptsize{$\pm$1 }}
                &&  7.8 \scriptsize{$\pm$0.1}
                 &  9.0 \scriptsize{$\pm$0.2}
                 & \textbf{26   \scriptsize{$\pm$1 }}
                 & 42   \scriptsize{$\pm$1 }
                 & 18   \scriptsize{$\pm$0 }
                 & 17   \scriptsize{$\pm$1 }
                 & 36   \scriptsize{$\pm$1 }
			\\
			\scriptsize \texttt{3}
			    & \texttt{PoolCMA}
                &  7.2 \scriptsize{$\pm$0.5}
                &  9.1 \scriptsize{$\pm$0.4}
                & 24   \scriptsize{$\pm$4 }
                & 41   \scriptsize{$\pm$2 }
                & 17   \scriptsize{$\pm$4 }
                & 16   \scriptsize{$\pm$2 }
                & 37   \scriptsize{$\pm$2 }
                &&  7.3 \scriptsize{$\pm$0.2}
                 &  9.0 \scriptsize{$\pm$0.3}
                 & 23   \scriptsize{$\pm$1 }
                 & 42   \scriptsize{$\pm$1 }
                 & 16   \scriptsize{$\pm$1 }
                 & 15   \scriptsize{$\pm$0 }
                 & 36   \scriptsize{$\pm$2 }
			\\
			\scriptsize \texttt{4}
			    & \texttt{PoolEndCMA}
                &  7.6 \scriptsize{$\pm$0.8}
                &  8.9 \scriptsize{$\pm$0.9}
                & 27   \scriptsize{$\pm$3 }
                & 42   \scriptsize{$\pm$3 }
                & 18   \scriptsize{$\pm$4 }
                & 17   \scriptsize{$\pm$2 }
                & 38   \scriptsize{$\pm$2 }
                &&  6.9 \scriptsize{$\pm$0.2}
                 &  \textbf{8.7 \scriptsize{$\pm$0.2}}
                 & 25   \scriptsize{$\pm$2 }
                 & \textbf{44   \scriptsize{$\pm$1 }}
                 & 18   \scriptsize{$\pm$1 }
                 & 16   \scriptsize{$\pm$1 }
                 & \textbf{38   \scriptsize{$\pm$2 }}
			\\
			\bottomrule
		\end{tabular}}
	\end{center}
	\caption{Cross-modal attention (CMA) model performance with \textit{unstructured memory} on \benchmarkce. 
	We compare tour-persistent memory (rows 2-4) against an episodic-memory baseline (row 1).
	Persisting in the environment over a tour does not improve performance on scenes not seen in training (Val-Unseen), unlike the behavior of \textit{semantic map models} (\tabref{tab:semantic_results}). 
	Metrics are $\bar{x} \pm \sigma_{\bar{x}}$ over 3 runs.
	}
	\label{tab:ivlnce_results}
\end{table*}


We present results on \benchmark\ and \benchmarkce\ below and center our discussion on key observations.

\paragraph{Evaluation Metrics}
We use \metric{t-nDTW} (\secref{sec:paradigm}) as our primary metric and report as a percentage.
We also include metrics standard to VLN and VLN-CE \cite{anderson2018evaluation, magalhaes2019effective} to describe average single-episode (\textit{episodic}) performance: trajectory length (\metric{TL}), navigation error (\metric{NE}), oracle success (\metric{OS}), normalized dynamic time warping (\metric{nDTW}), success rate (\metric{SR}), and success weighted by inverse path length (\metric{SPL}).

\subsection{Unstructured Memory in \benchmark\ and \benchmarkce}
\paragraph{Naive extensions of unstructured memory fail to exploit tour information.} Across both  \benchmark\  and \benchmarkce, we find that the simple extensions of unstructured memory explored in this work fail to improve (and often hurt) tour performance compared to purely episodic agents. 

\emph{\benchmark.} \tabref{tab:ivln_results} shows results of HAMT and TourHAMT variations in the discrete \benchmark\ setting. We find the episodic HAMT model to be a strong baseline even without any tour memory; our attempts to add tour memory significantly degraded performance. Simply extending the memory reduces \metric{t-NDTW} by a factor of two from HAMT (row 1 vs.~5). Finetuning the history encoding for this extended memory setting (row 4), adding a \texttt{PREV} token to separate memory from prior episodes (row 3), and applying inflection weighting \cite{wijmans2019emboied} (row 2) collectively regain 4\% \metric{t-nDTW} but still falls short of the HAMT baseline (row 1 vs.~2). As our experiments start from a pretrained HAMT model, we speculate these results are due to the history encoder coping poorly with a distribution shift in its inputs compared to the pretraining tasks and that direct finetuning is insufficient to correct for this. 

\emph{\benchmarkce.} \tabref{tab:ivlnce_results} shows results of the CMA-based models in the continuous \benchmarkce\ setting. We find that naive extension of unstructured memory reduces performance (row 1 vs.~2), but pooling versions recover tour performance (rows 3-4). Notably, all models augmented with tour memory yield reductions in episodic metrics even when tour metrics are comparable. We draw attention to the \texttt{TourCMA} model performance on Val-Seen (row 2, left) which exhibits stronger performance than the baseline CMA. 
This result suggests tour-memory allows more overfitting than episodic memory.

We find that more sophisticated memory structures or training methods will be needed to capitalize on \paradigm.
We explore one such memory structure.



\subsection{Semantic Map Memory in \benchmarkce}
\label{sec:ivlnce_results}

\setlength{\lightrulewidth}{0.08em}
\setlength{\tabcolsep}{.5em}
\begin{table*}[t]
    \setlength{\aboverulesep}{0pt}
    \setlength{\belowrulesep}{0pt}
    \renewcommand{\arraystretch}{1.15}
	\begin{center}
	\resizebox{0.8\textwidth}{!}{
		\begin{tabular}{clll c cccccs c ccccccs}
			\toprule
            & & \multicolumn{2}{c}{\scriptsize \shortstack{Map Construction}}

			& \multicolumn{7}{c}{\scriptsize\textbf{Val-Seen}}
		   && \multicolumn{7}{c}{\scriptsize\textbf{Val-Unseen}}
            \\
			\cmidrule{5-11}
			\cmidrule{13-19}

            \scriptsize \shortstack{\#} 
            & \scriptsize \shortstack{Semantics}
            & {\scriptsize Train}
            & {\scriptsize Eval}

			& \scriptsize\textbf{\texttt{TL}}
			& \scriptsize\textbf{\texttt{NE}}~$\downarrow$
			& \scriptsize\textbf{\texttt{OS}}~$\uparrow$
			& \scriptsize\textbf{\texttt{nDTW}}~$\uparrow$
			& \scriptsize\textbf{\texttt{SR}}~$\uparrow$
			& \scriptsize\textbf{\texttt{SPL}}~$\uparrow$
			& \scriptsize\textbf{\texttt{t-nDTW}}~$\uparrow$
			&
			& \scriptsize\textbf{\texttt{TL}}
			& \scriptsize\textbf{\texttt{NE}}~$\downarrow$
			& \scriptsize\textbf{\texttt{OS}}~$\uparrow$
			& \scriptsize\textbf{\texttt{nDTW}}~$\uparrow$
			& \scriptsize\textbf{\texttt{SR}}~$\uparrow$
			& \scriptsize\textbf{\texttt{SPL}}~$\uparrow$
			& \scriptsize\textbf{\texttt{t-nDTW}}~$\uparrow$
			\\

			\midrule
			\scriptsize \texttt{1}
                & \multirow{7}{*}{\shortstack[l]{Ground- \\ Truth}} 
                & \multirow{3}{*}{Ep.} 
                & Ep. 
                & 10.6 & 6.3 & 51 & 54 & 34 & 32 & 49 
               &&  9.8 & 6.9 & 42 & 50 & 29 & 26 & 42 
			\\
			\scriptsize \texttt{2}
                & 
                & 
                & It. 
                & 10.4 & 6.3 & 50 & 54 & 34 & 31 & 50 
               && 9.4 & 7.0 & 41 & 50 & 29 & 27 & 43 
			\\
			\scriptsize \texttt{3}
                & 
                & 
                & Kn. 
                & 10.1 & 6.2 & 50 & 55 & 34 & 32 & 50 
               && 9.5 & 6.9 & 40 & 50 & 29 & 26 & 43 
			\\
			\cmidrule{3-19}
			\scriptsize \texttt{4}
                & 
                & \multirow{3}{*}{It.} 
                & Ep. 
                & \hphantom{0}9.4 & 6.8 & 45 & 54 & 36 & 34 & 51 
               && 8.7 & 7.1 & 40 & 52 & 32 & 30 & 44 
			\\
			\scriptsize \texttt{5}
                & 
                & 
                & It. 
                & \hphantom{0}9.5 & 6.3 & 52 & \textbf{58} & 41 & \textbf{39} & \textbf{54} 
               && 8.5 & \textbf{6.7} & 43 & \textbf{54} & \textbf{36} & \textbf{33} & 48 
			\\
			\scriptsize \texttt{6}
                & 
                & 
                & Kn. 
                & \hphantom{0}9.4 & 6.2 & 51 & \textbf{58} & \textbf{42} & \textbf{39} & \textbf{54} 
               && 8.6 & \textbf{6.7} & 43 & \textbf{54} & 34 & 32 & \textbf{49} 
			\\
			\cmidrule{3-19}
			\scriptsize \texttt{7}
                & 
                & \multirow{3}{*}{Kn.} 
                & Ep. 
                & 10.0 & 7.3 & 43 & 50 & 32 & 29 & 46 
               && 9.4 & 7.7 & 37 & 48 & 29 & 27 & 40 
			\\
			\scriptsize \texttt{8}
                & 
                & 
                & It. 
                & \hphantom{0}9.9 & 6.4 & 49 & 55 & 37 & 34 & 51 
               && 9.1 & 6.8 & \textbf{45} & 53 & 34 & 31 & 46 
			\\
			\scriptsize \texttt{9}
                & 
                & 
                & Kn. 
                & \hphantom{0}9.9 & \textbf{6.1} & \textbf{54} & \textbf{58} & 41 & \textbf{39} & 51 
               && 9.2 & \textbf{6.7} & 44 & 53 & 34 & 32 & 46 
			\\[1ex]
			\midrule
			\scriptsize \rule{0pt}{4.5ex}\texttt{10}
                & \multirow{7}{*}{\shortstack[l]{Inferred}} 
                & \multirow{3}{*}{Ep.} 
                & Ep. 
                & 10.2 & 6.6 & 51 & 54 & 35 & 32 & 50 
               && 9.8 & 7.2 & 43 & 50 & 33 & 30 & 43 
			\\
			\scriptsize \texttt{11}
                & 
                & 
                & It. 
                & 10.1 & 6.6 & 49 & 54 & 36 & 33 & 51 
               && 9.3 & 7.5 & 39 & 49 & 29 & 27 & 42 
			\\
			\scriptsize \texttt{12}
                & 
                & 
                & Kn. 
                & 10.0 & 6.9 & 49 & 53 & 34 & 32 & 50 
               && 9.3 & 7.3 & 41 & 50 & 29 & 27 & 43 
			\\
			\cmidrule{3-19}
			\scriptsize \texttt{13}
                & 
                & \multirow{3}{*}{It.} 
                & Ep. 
                &  \hphantom{0}9.5 & 6.9 & 43 & 53 & 34 & 31 & 48 
               &&  8.8 & 7.3 & 40 & 51 & 31 & 29 & 44 
			\\
			\scriptsize \texttt{14}
                & 
                & 
                & It. 
                &  \hphantom{0}9.4 & 6.4 & 48 & 56 & 39 & 36 & 52 
               &&  8.5 & \textbf{6.8} & \textbf{44} & \textbf{54} & \textbf{35} & \textbf{32} & \textbf{47} 
			\\
			\scriptsize \texttt{15}
                & 
                & 
                & Kn. 
                &  \hphantom{0}9.3 & 6.4 & 51 & 56 & 38 & 36 & 52 
               &&  8.6 & 6.9 & 43 & \textbf{54} & 34 & 31 & 46 
			\\
			\cmidrule{3-19}
			\scriptsize \texttt{16}
                & 
                & \multirow{3}{*}{Kn.} 
                & Ep. 
                & 10.2 & 6.7 & 50 & 53 & 35 & 32 & 50 
               &&  9.6 & 7.5 & 39 & 49 & 28 & 25 & 41 
			\\
			\scriptsize \texttt{17}
                & 
                & 
                & It. 
                & 10.0 & \textbf{6.1} & \textbf{57} & \textbf{57} & \textbf{40} & 36 & \textbf{54} 
               &&  9.4 & 7.1 & \textbf{44} & 51 & 30 & 27 & 43 
			\\
			\scriptsize \texttt{18}
                & 
                & 
                & Kn. 
                & 10.1 & 6.2 & 55 & \textbf{57} & \textbf{40} & \textbf{37} & 53 
               &&  9.4 & 7.2 & 43 & 51 & 31 & 28 & 44 
			\\
			\bottomrule
		\end{tabular}}
	\end{center}
	\caption{Performance of \model{MAP-CMA} agents in \benchmarkce. We consider resetting maps each episode (\textit{Ep.}), constructing maps throughout tours (\textit{It.}), and knowing maps from the start (\textit{Kn.}). We construct maps from ground-truth semantics in rows 1-9 and infer semantics from RedNet \cite{jiang2018rednet} in rows 10-18. We use bolding to highlight best scores in ground-truth and inferred semantics separately. Iteratively constructing tour maps leads to better performance than using single-episode maps.}
	\label{tab:semantic_results}
\end{table*}

\label{sec:semantic_results}

Agents in \benchmarkce\ complete long tours (average of 50 R2R episodes per tour) using low-level actions, resulting in some tours requiring over two thousand actions. 
Storing tour memory as an unstructured vector that is updated per-step (as in the \model{TourCMA}/\model{PoolCMA}/\model{PoolEndCMA} agents) does not effectively remember the environment, and Transformer-based agents may face adverse scaling. 
We instead consider a \textit{structured} memory in the form of a metric map of semantics and occupancy. 
We present results varying: the map source \{Ground Truth, Inferred via RedNet\}; mapping procedure during training \{Episodic, Iterative, Known\}; and mapping procedure during evaluation \{Episodic, Iterative, Known\} on \benchmarkce\ in \tabref{tab:semantic_results} for a total of 18 settings.\footnote{`Ground-Truth' agents have access to oracle semantic and occupancy information and evaluations performed with the `Known' mapping procedure assume a pre-explored environment. 
These methods do not constitute valid submissions to the \benchmarkce\ leaderboard, but are provided for analysis.}

\paragraph{Map-based memory can leverage prior experience.} 
Across all map sources and map procedures during training, we find agents perform better with iteratively-updated maps persisting across tours (Eval: It.) than with episodic maps reset between episodes (Eval: Ep.). 
This difference confirms that information gathered from previous episodes in a tour can benefit following novel instructions along novel paths.
This effect is strongest when the agent was trained with iteratively-updated maps (Train: It.) with \metric{t-nDTW} improving by 3-4\% for ground truth (row 4 vs.~5) and RedNet-inferred maps (row 13 vs.~14). 
Models trained on episodic maps find the least benefit from iteratively-updated maps (row 1 vs.~2 and 10 vs.~11), suggesting utilizing semantic map memory beyond the current episode is a learned skill.

\paragraph{Map-structured memories may be better suited to \benchmarkce.} Across all settings, we find map-based agents outperform the CMA models from \tabref{tab:ivlnce_results}, with best performing iterative agents achieving +9 \metric{t-nDTW} (24\%). However, differences in training procedure may account for some of this gap. Map-based agents underwent DAgger-based fine-tuning but CMA agents did not. In prior work in VLN-CE, DAgger training CMA improved episodic nDTW by 1-5 points \cite{krantz_vlnce_2020}.

\paragraph{Ground truth and inferred semantics perform similarly.}
Inferring semantics with RedNet (rows 10-18) leads to a small, consistent drop in performance when compared to using ground-truth semantics (rows 1-9).
However, the best performing agent (row 14) sees only a 1 point drop (row 5) in \metric{t-nDTW}.
This limited sensitivity may be due to agents not making full use of the semantic labels or that the scope of labels is not broad enough for highly-performant agents.

\paragraph{Known maps fail to outperform iterative maps.} Agents trained and evaluated with known maps fail to outperform those trained and evaluated with iterative maps for both ground-truth (row 9 \vs 5) and inferred (row 18 \vs 14) semantics, yet known maps outperform training on episodic maps (rows 1, 10). Models trained with iterative maps may benefit from exposure to the divide between explored and unexplored regions. The relatively low performance of known maps points to the open question of how to effectively encode and decode scene perception for downstream reasoning.
\benchmarkce\ can be a fruitful arena for such a study.

\section{Conclusions}
\label{sec:conclusion}

We define \paradigmfull\ (\paradigm), a paradigm for studying how language-guided agents persisting in a scene, like robots in a home, can utilize past experience to follow instructions.
We create the \benchmark\ and \benchmarkce\ benchmarks to study discrete and continuous navigation across \textit{tours} comprising many single-instruction \textit{episodes}.
Initial models for both benchmarks show that extending unstructured latent memory beyond episode scope is insufficient to generalize to tours, but agents that build explicit maps benefit from environment persistence.

\paragraph{Limitations}
Our benchmarks are limited to English instructions and the indoor spaces are largely lavish, staged homes and offices.
Deployed assistive robots performing navigation should respond to more than English, and should be able to navigate cluttered, realistic home environments.
Such biases in a benchmark serve the needs of English-speaking, able-bodied folks as a ``default,'' and will hinder such long term goals in spaces of human-robot interaction and accessibility.

\paragraph{Future Work}
Even with the benefit of tours, \model{MAP-CMA} lags far behind human performance on independent episodes, with an episodic SPL of 32 in IR2R-CE Val-Unseen vs. 76 for humans on R2R Test. We anticipate progress may require improved methods for grounding natural language into maps and actions; iteratively constructed maps that are more accurate, flexible and expressive; and improved methods for transfer learning or generating synthetic data.

\paragraph{Acknowledgments and Disclosure of Funding}
The Oregon State effort is supported in part by the DARPA Machine Common Sense program. 
The U of Michigan effort is supported in part by the NSF COVE initiative under Grant 1628987 and Google. 
The views and conclusions contained herein are those of the authors and should not be interpreted as representing the official policies or endorsements, either expressed or implied, of the US Government or any sponsor.

{\small
\bibliographystyle{ieee_fullname}
\bibliography{egbib}
}

\clearpage

\appendix
\section*{Supplementary Material}

\section{Implementation Details}

\paragraph{IVLN Experiments}
We used a batch size of 8, a learning rate of 1e-5 and trained with the AdamW optimizer for all the IVLN models.
The maximum number of steps was 15 per episode. 
In training, we chose a dropout rate of 0.5 and a environment dropout rate of 0.4.
For the baseline HAMT model, we trained for a total of 100k iterations and evaluated every 2k iterations. 
For the \thamt models, we trained for a total of 5k iterations and evaluated every 200 iterations.
Note that \thamt models were trained with batches of tours and the HAMT model was trained with batches of episodes.
We trained all models with teacher-forcing and updated gradients per episode.
Thus, \thamt models did not train with fewer gradient updates than the HAMT model. 
We ran IVLN experiments on Quadro RTX 6000 GPUs. The fine-tuning of the baseline HAMT model took 12 GPU hours and the training of each \thamt model took 70 GPU hours. Altogether, a total of 876 GPU hours were used for the IVLN experiments.

\paragraph{IVLN-CE Experiments} 
We used a batch size of 5 for all training phases. For both teacher-forcing and DAgger training phases, we used a learning rate of 2.5e-4 with the Adam optimizer. For DAgger training, we performed 10 iterations with 4 epochs of training per iteration. We collected 5000 rollouts per iteration. We took the oracle action with probability $\beta = 0.5^{n+1}$ where $n$ is the index of the current iteration and took the agent action otherwise. 
We evaluated the model after each epoch of training. During inference the model takes the argmax of the predicted action distribution.
IVLN-CE experiments were run on Tesla V100 GPUs. Each unstructured latent memory model required 24 GPU hours to train and each semantic map model required 120 GPU hours to train. Altogether, 2016 total GPU hours
were used for training and evaluation to support these experiments.

\section{Evaluation Server Details}

As mentioned in \secref{sec:intro}, we will release an evaluation server and public leaderboard to standardize and benchmark progress on \paradigm. Like existing evaluation servers for VLN\footnote{R2R server: \href{https://eval.ai/web/challenges/challenge-page/97}{eval.ai/web/challenges/challenge-page/97}} 
and VLN-CE,\footnote{R2R-CE server: \href{https://eval.ai/web/challenges/challenge-page/719}{eval.ai/web/challenges/challenge-page/719}} we will establish \benchmark\ and \benchmarkce\ evaluation servers that accept and score prediction files. Participants will run inference of their models locally on a Test split. The prediction file will be scored on the evaluation server and the results will be added to a public leaderboard.

In existing evaluation servers of the Room-to-Room dataset, the target paths of the Test split are kept private to preserve the integrity of the split. However, in \paradigm, the \textit{oracle navigation} phase involves conveying the agent to the target goal location at the end of each episode in a tour. Running model inference in such a setting requires access to the target path. Given this concern, we modify the \paradigm\ paradigm for leaderboard evaluation as follows:

\begin{enumerate}
	\item The \textit{oracle navigation} phase is modified to only convey the agent to the start of the next episode, rather than first to the target goal of the current episode. 
	\item Episodes in a tour are ordered such the distance between the start locations of sequential episodes is minimized (tip-to-tip). Ordering episodes tip-to-tail as originally presented for \paradigm\ would compromise the target goal locations with high certainty.
\end{enumerate}

\begin{figure}[t]
    \centering
	\resizebox{\columnwidth}{!}{
        \includegraphics[width=\textwidth]{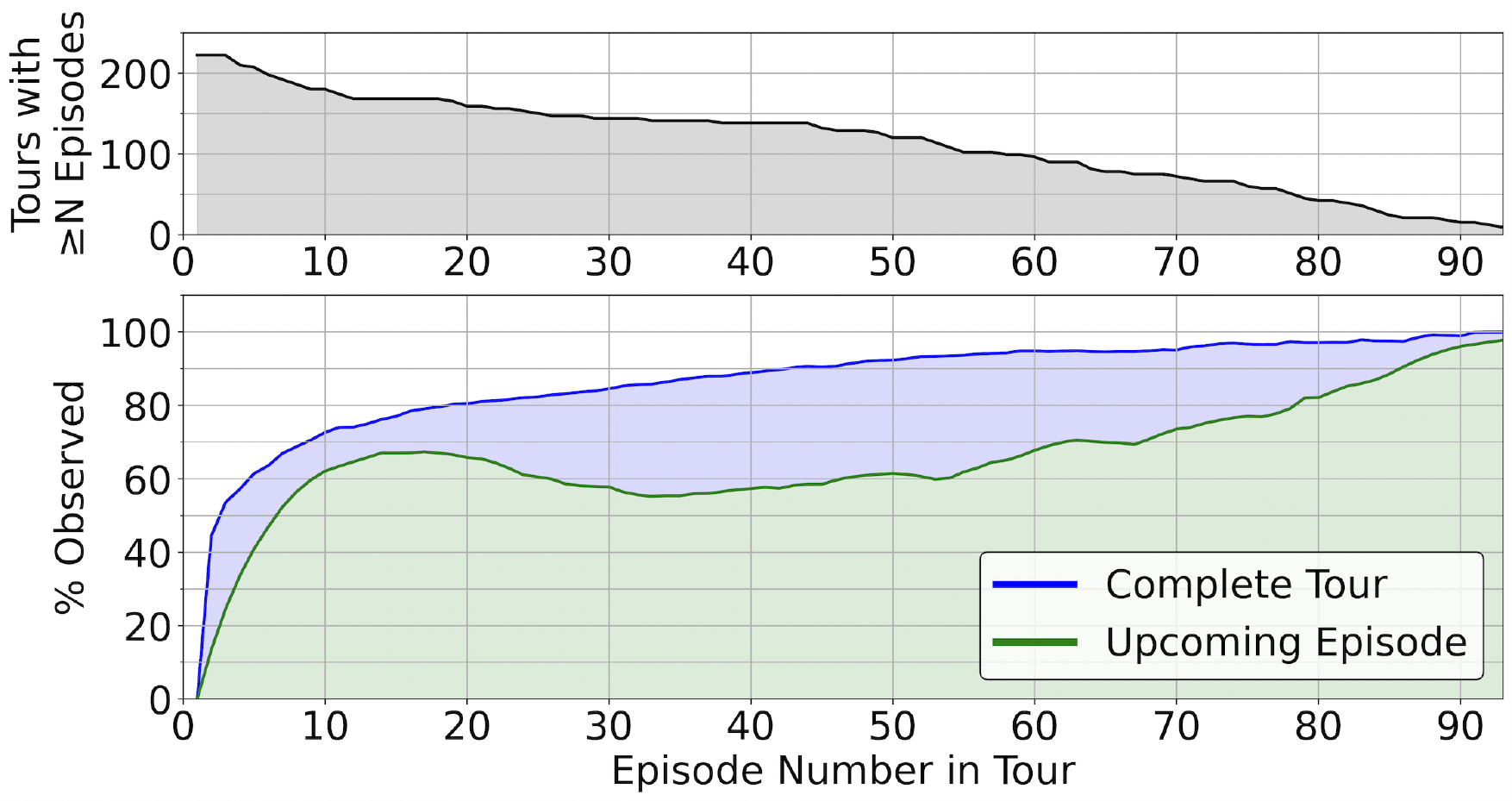}
    }
    \caption{We repeat \figref{fig:tour_obs} for the \benchmarkce{} Train split regenerated using Test split modifications. This demonstrates that during a tour, the percentage already observed of both the \textcolor{pltgreen}{Upcoming Episode} and the \textcolor{blue}{Complete Tour} are similar to the observation coverage experienced while following the tours proposed in the main paper.}
    \label{fig:tour_obs_tip_to_tip}
\end{figure}

These two modifications ensure that the privacy of the Test split target paths of Room-to-Room are preserved. A second concern is the domain gap between the Test split and rest of the \paradigm\ splits induced by the modifications above. We regenerate the Train and validation splits of \benchmarkce\ with these modifications. In \figref{fig:tour_obs_tip_to_tip}, we show the observation coverage of both upcoming episodes and complete tours of the regenerated Train split for direct comparison to \figref{fig:tour_obs}. We find that these modifications result in a slightly lower AUC of \textcolor{blue}{Complete Tour} coverage while \textcolor{pltgreen}{Upcoming Episode} coverage is mostly similar. Finally, we evaluate our best \model{Map-CMA} model (\tabref{tab:semantic_results} row 14) against the regenerated \benchmarkce\ Val-Unseen split. Performance metrics are similar to those reported in \tabref{tab:semantic_results}: 48 \metric{t-nDTW} (+1), 36 \metric{SR} (+1), and 33 \metric{SPL} (+1).
Thus, we conclude these modifications for Test set evaluation on a public leaderboard ensure the privacy of the R2R test paths while preserving the core \paradigm\ evaluation challenges.

\section{Illustrative Figures}

\begin{figure*}[htp]
    \centering
    \includegraphics[width=0.86\textwidth]{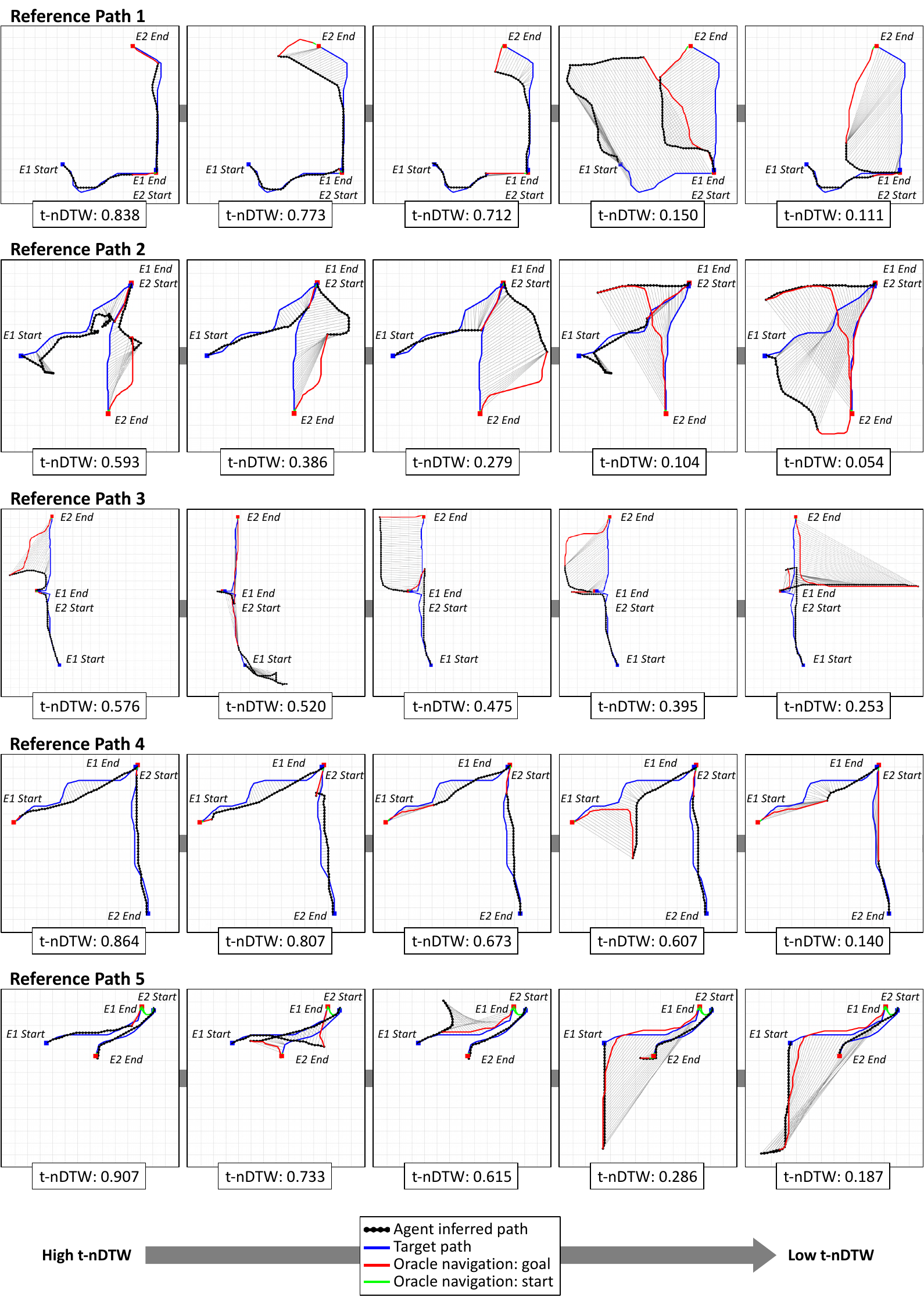}
    \caption{
        Ranked sample evaluations of \metric{t-nDTW} scores applied to five reference paths. 
        While \metric{t-nDTW} evaluates the entire length of a tour, here we evaluate 2-episode sub-tours for illustration.
        }
    \label{fig:ndtw_all_examples}
\end{figure*}

\begin{figure*}[htp]
    \centering
    \includegraphics[width=0.85\textwidth]{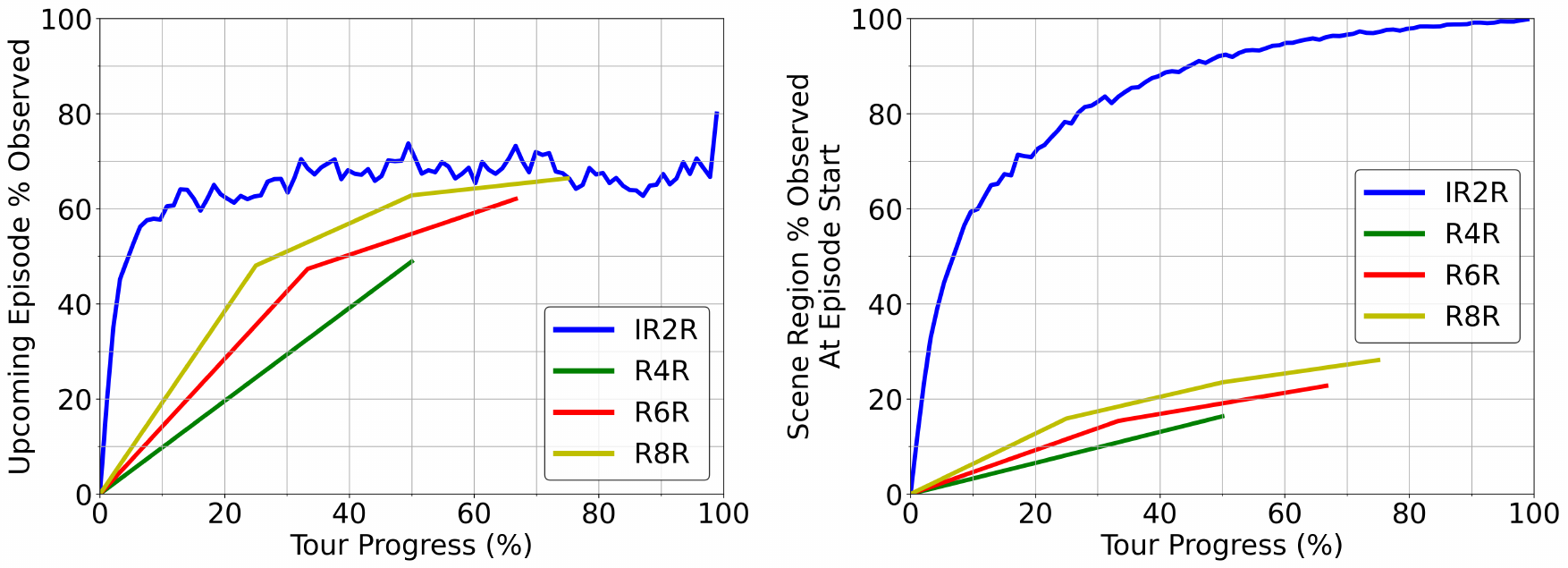}
    \caption{
    We compare \benchmarkce\ against the continuous environment versions of R4R, R6R, and R8R. 
    Room-for-Room (R4R) is a benchmark in VLN consisting of long-distance instruction-following episodes synthesized from the Room-to-Room (R2R) dataset by joining two adjacent paths (and consequently instructions) tip-to-tail \cite{jain-etal-2019-stay}. Joining 3 paths tip-to-tail is known as R6R and joining 4 paths tip-to-tail is known as R8R \cite{babywalk}. 
    In this comparison, at the start of each episode we measure what percentage of that episode's target path has been observed earlier in the tour (left). Since the tours of \benchmarkce\ are significantly longer on average (100 episodes \vs 2, 3, or 4), the agent has a higher percentage of prior observability when tasked with a new instruction to follow. Also at the start of each episode, we measure what percentage of the observable scene region has been observed (right). Agents performing tours in \benchmarkce\ smoothly gain observation coverage of the scene region, eventually observing the entire space. On the other hand, agents performing tours in R8R average less than 30\% observability coverage by the beginning of the final episode. The significantly higher and iteratively-procured episode coverage and scene region coverage in \paradigm\ provides a realistic and rich signal for navigation planning that has yet to be featured in any other VLN benchmark.
    }
    \label{fig:cov_comparison}
\end{figure*}

\begin{figure*}[htp]
    \centering
    \includegraphics[width=0.65\textwidth]{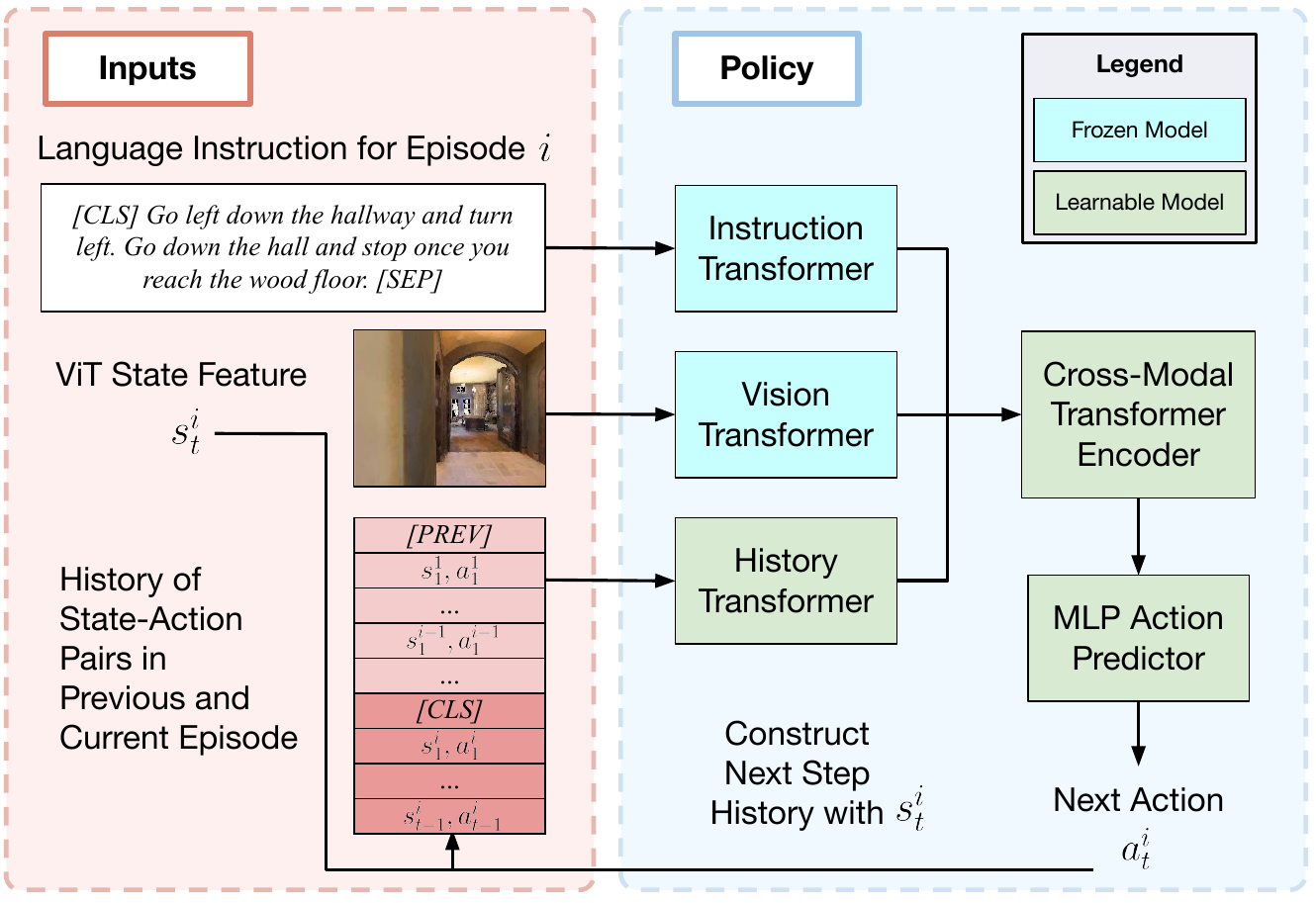}
    \caption{The TourHAMT model architecture. Unlike the original HAMT model~\cite{chen2021history}, we unfreeze the History Transformer module, and append the previous episodes' state-action pairs as history (light red top part of the history box) on top of the current episode's history (dark red bottom part of the history box). The state-action history is empty at the beginning of the tour and accumulates as the number of experienced episodes grows.}
    \label{fig:model_thamt}
\end{figure*}

\begin{figure*}[htp]
    \centering
    \includegraphics[width=0.83\textwidth]{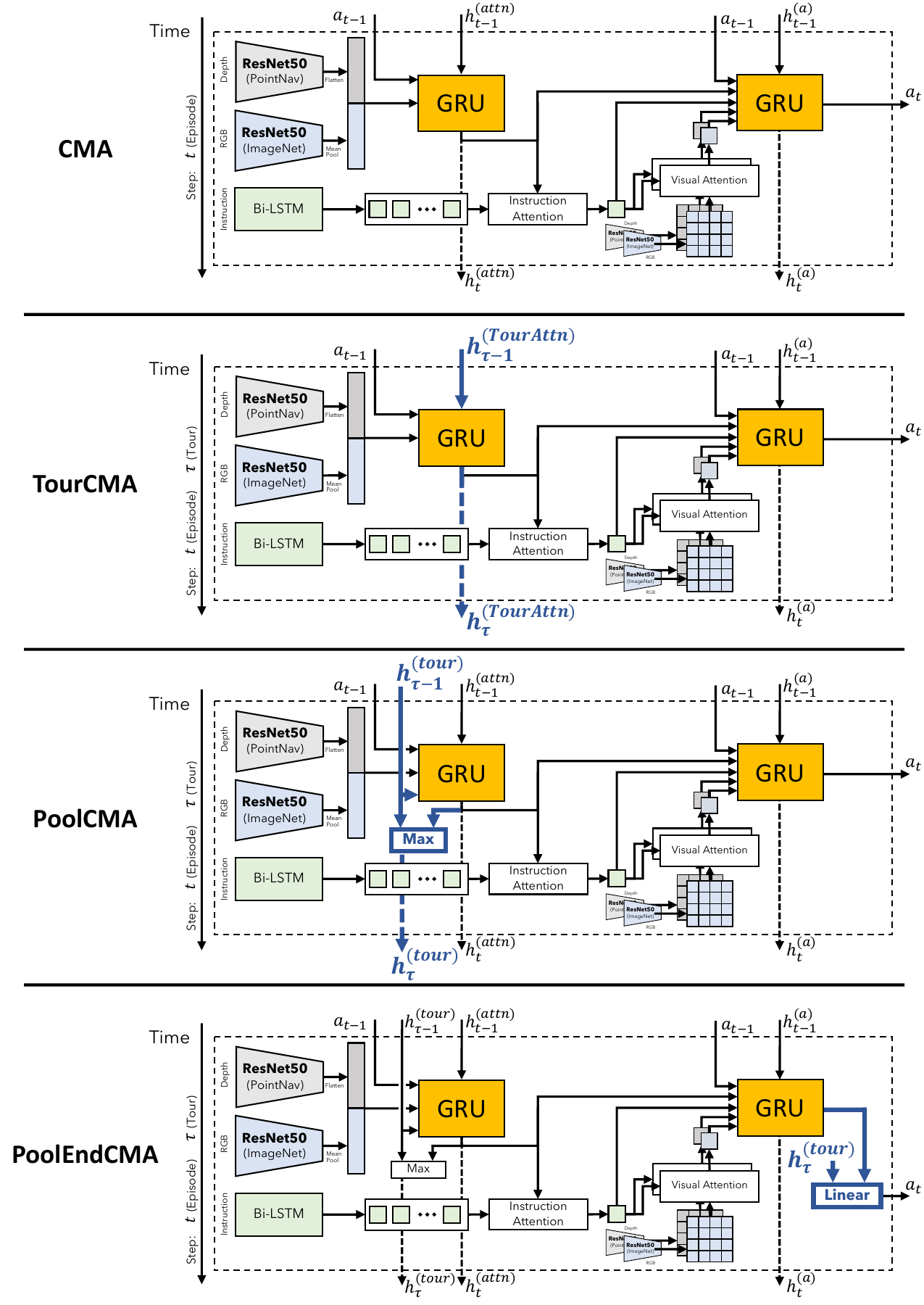}
    \caption{Architectures of unstructured memory models in \benchmarkce. Episode-based \model{CMA} as defined in \cite{krantz_vlnce_2020} is shown on top. Tour-adapted versions are shown below with $t$ indicating a step in the current episode and $\tau$ indicating a step in the current tour. Memory structures with episode steps, $h_t$, are reset for new episodes, whereas memory structures with tours steps, $h_{\tau}$, are reset for new tours. Changes from the model immediately above are shown in bold in dark blue.
    }
    \label{fig:model_latent}
\end{figure*}

Figures~\ref{fig:ndtw_all_examples},~\ref{fig:cov_comparison},~\ref{fig:model_thamt}, and~\ref{fig:model_latent} further illustrate \paradigm\ and proposed, initial models.

\end{document}